%% file: main.tex
\documentclass[10pt,twocolumn,letterpaper]{article}

\usepackage{cvpr}              % To produce the CAMERA-READY version
\definecolor{cvprblue}{rgb}{0.21,0.49,0.74}
\usepackage[pagebackref,breaklinks,colorlinks,allcolors=cvprblue]{hyperref}

\def\paperID{11660} % *** Enter the Paper ID here
\def\confName{CVPR}
\def\confYear{2025}

\title{Bayesian Prompt Flow Learning for Zero-Shot Anomaly Detection}

\usepackage{algorithm}
\usepackage{algorithmic}
\author{
	Zhen Qu$^{1,2}$ \quad Xian Tao$^{1,2,4}\textsuperscript{\Letter}$ \quad Xinyi Gong$^{3}$ \quad Shichen Qu$^{1,2}$ \quad
	Qiyu Chen$^{1,2}$ \and
	Zhengtao Zhang$^{1,2,4}$ \quad Xingang Wang$^{1,2,5}$ \quad Guiguang Ding$^6$ \\
	$^1$Institute of Automation, Chinese Academy of Sciences \\ 
	$^2$School of Artificial Intelligence, University of Chinese Academy of Sciences \quad $^3$HDU\\
	$^4$Casivision \quad $^5$Luoyang Institute for Robot and Intelligent Equipment \quad $^6$Tsinghua University \\
	{\tt\small \{quzhen2022, xiantao2013, qushichen2023, chenqiyu2021, xingang.wang, zhengtao.zhang\}@ia.ac.cn} \\
	{\tt\small gongxinyi@hdu.edu.cn} \quad {\tt\small dinggg@tsinghua.edu.cn}
}
\usepackage{colortbl}
\usepackage{multirow}
\usepackage{wrapfig}
\usepackage{tabularx}
\usepackage{makecell}
\usepackage{enumitem}
\usepackage{marvosym}
\usepackage[accsupp]{axessibility}  % Improves PDF readability for those with disabilities.
\usepackage{float} % 加载 float 宏包
\usepackage{pifont}
\definecolor{lightgreen}{RGB}{240,255,255}
\definecolor{darkred}{RGB}{180,0,0} % 定义暗红色
\definecolor{lightblue1}{RGB}{100, 149, 237} % Deeper blue
\definecolor{lightgreen1}{RGB}{34, 139, 34} % Deeper green
\definecolor{lightorange1}{RGB}{255, 165, 0} % Deeper orange

\begin{document}
	\maketitle
	\input{sec/0_abstract}    
	\input{sec/1_intro}

	\input{sec/2_related_work}

	\input{sec/3_method}

	\input{sec/4_experiments}

	\input{sec/5_discussion}
	{
		\small
		\bibliographystyle{ieeenat_fullname}
		\bibliography{main}
	}
	
	\input{sec/X_suppl}
	
\end{document}

%% file: sec/0_abstract.tex
\begin{abstract}
Recently, vision-language models (e.g. CLIP) have demonstrated remarkable performance in zero-shot anomaly detection (ZSAD). By leveraging auxiliary data during training, these models can directly perform cross-category anomaly detection on target datasets, such as detecting defects on industrial product surfaces or identifying tumors in organ tissues. Existing approaches typically construct text prompts through either manual design or the optimization of learnable prompt vectors. However, these methods face several challenges: 1) handcrafted prompts require extensive expert knowledge and trial-and-error; 2) single-form learnable prompts struggle to capture complex anomaly semantics; and 3) an unconstrained prompt space limits generalization to unseen categories. To address these issues, we propose Bayesian Prompt Flow Learning (Bayes-PFL), which models the prompt space as a learnable probability distribution from a Bayesian perspective. Specifically, a prompt flow module is designed to learn both image-specific and image-agnostic distributions, which are jointly utilized to regularize the text prompt space and improve the model's generalization on unseen categories. These learned distributions are then sampled to generate diverse text prompts, effectively covering the prompt space. Additionally, a residual cross-model attention (RCA) module is introduced to better align dynamic text embeddings with fine-grained image features. Extensive experiments on 15 industrial and medical datasets demonstrate our method's superior performance. The code is available at \url{https://github.com/xiaozhen228/Bayes-PFL}.
\end{abstract}

%% file: sec/1_intro.tex
\section{Introduction}
\label{sec:intro}
\begin{figure}[tb]
	\centering
	\includegraphics[width=0.95\columnwidth]{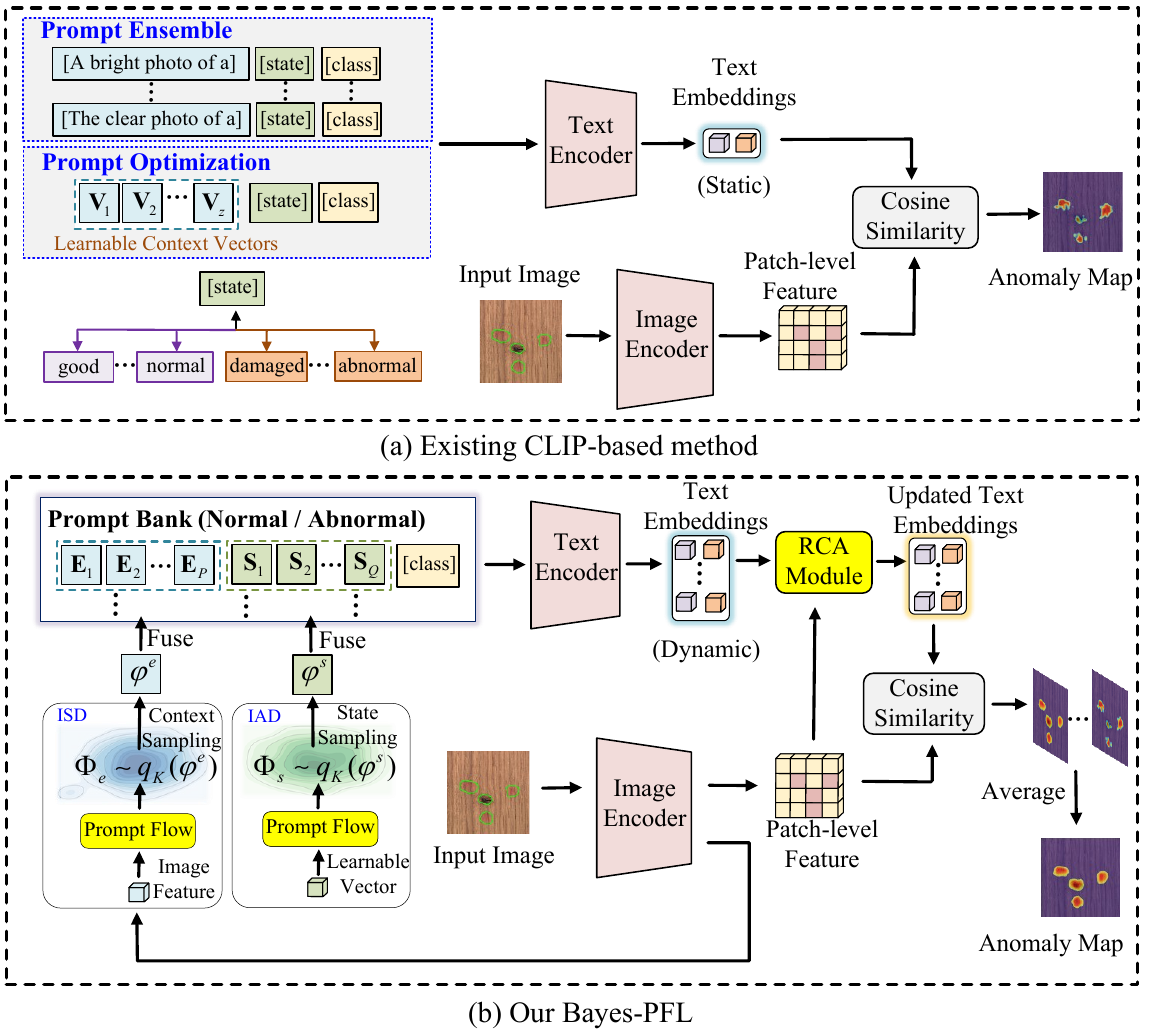}
	\caption{Comparison of existing CLIP-based method and our Bayes-PFL. 
		    Our Bayes-PFL learns the distributions of context and state words within textual prompts. The proposed RCA module is further to enhance the alignment between dynamic text embeddings and patch-level features through cross-modal interaction.}
	\label{Fig1}
\end{figure}
In the fields of industrial defect detection \cite{industry1, yolo10, WinCLIP, investigating} and medical image analysis \cite{medical1, ISIC, medical2, AdaCLIP}, the cold-start problem is a major challenge, characterized by an insufficient amount of labeled data for supervised training on new object categories. An effective solution is zero-shot anomaly detection (ZSAD), which directly detects anomalies on new object surface after being trained on auxiliary data. However, significant variations in background features, anomaly types, and visual appearances across different products and organs make it difficult to achieve robust generalization.
\par
 Vision-language models (VLMs), such as CLIP \cite{CLIP} and ALIGN \cite{ALIGN}, which are trained on large-scale image-text pairs through contrastive representation learning, have shown significant potential for ZSAD. As a representative VLM, CLIP \cite{CLIP} has been widely adopted in previous works \cite{WinCLIP, AnomalyCLIP, AdaCLIP}. A key challenge for CLIP-based ZSAD methods is the effective design of text prompts. As illustrated in Figure \ref{Fig1}(a), a typical textual prompt consists of three components: \textcolor{lightblue1}{\textit{context words}}, \textcolor{lightgreen1}{\textit{state words}}, and \textcolor{lightorange1}{\textit{class}}. They respectively represent a general description of the object (e.g. background), the attributes indicating whether it is normal or anomalous (e.g. good / damaged), and the object category (e.g. wood). Existing text prompt design methods can be broadly categorized into two types: prompt ensemble-based and prompt optimization-based. Prompt ensemble-based methods, such as WinCLIP \cite{WinCLIP}, APRIL-GAN \cite{VAND}, and CLIP-AD \cite{CLIPAD}, create handcrafted prompt templates and embed various context and state words into these templates to generate a large number of textual prompts. However, the number of context and state words is limited and the design relies on expert knowledge. In contrast, prompt optimization-based methods, such as AnomalyCLIP \cite{AnomalyCLIP} and AdaCLIP \cite{AdaCLIP}, replace the context words in the prompts with learnable vectors or directly insert them into the encoder. However, two critical issues are often overlooked: 1) The design of learnable prompts is overly simplistic, which makes it difficult to capture complex context semantics; 2) The learnable prompt space is not properly constrained, hindering the generalization of learnable text prompts to unseen categories.
\par 
 To address these challenges, we propose a Bayesian Prompt Flow Learning (Bayes-PFL) strategy based on CLIP for the ZSAD task. As depicted in Figure \ref{Fig1}(b), our approach learns a probability distribution over textual prompts from a Bayesian inference perspective, aiming to improve the ZSAD performance through two key aspects: 1) regularizing the context and state text spaces separately through the design of image-specific distribution (ISD) and image-agnostic distribution (IAD), ensuring that the learned prompts generalize effectively to novel categories, and 2) enhancing the coverage of the prompt space by sampling prompts from the learned probability distribution.
 \par 
 First, we construct two-class prompt banks to store learnable textual prompts for normal and abnormal cases. Then, a prompt flow module is designed to learn the distribution of context and state words within the text prompts. Specifically, to introduce rich visual semantics into the text prompts, ISD is learned through prompt flow to model the context words, dynamically adapting based on the input image. To encode unified normal and abnormal semantics, IAD is established for the state words, remaining static and conditioned on a learnable free vector. Finally, multiple textual prompts sampled from the learned distribution are fused with the prompt banks to generate enriched text prompts. Moreover, to align the sampled text embeddings with fine-grained image features, we introduce a residual cross-modal attention (RCA) module to enhance the interaction between modalities. Unlike existing methods \cite{VAND,CLIPAD,AdaCLIP} that directly align text with patch-level features as shown in Figure \ref{Fig1}(a), our RCA module updates dynamic text embeddings using fine-grained image information to improve the ZSAD performance. The results generated from alignment of multiple updated text embeddings and patch-level feature are integrated to produce the final anomaly map.
\par 
The contributions of this work are summarized as follows: 1) We propose a novel Bayesian Prompt Flow Learning method, Bayes-PFL, built on CLIP to address the ZSAD problem. By distribution sampling and distribution regularization of the text prompt space, Bayes-PFL enhances the ZSAD performance on novel categories; 2) The prompt flow module is designed to learn ISD and IAD, which model the context and state semantics in text prompts, respectively. Additionally, the RCA module is introduced to improve the alignment between fine-grained features and dynamic text prompts through cross-modal interactions; 3) After training on auxiliary datasets, our method can directly detect defects / lesions on novel products / organs without requiring target training data. Extensive experiments across 15 industrial and medical datasets demonstrate that Bayes-PFL achieves state-of-the-art (SOTA) performance in ZSAD.
\par

%% file: sec/2_related_work.tex
\section{Related Works}
\label{Related_work}
	\begin{figure*}[t]
	\centering
	\includegraphics[width=1.9\columnwidth]{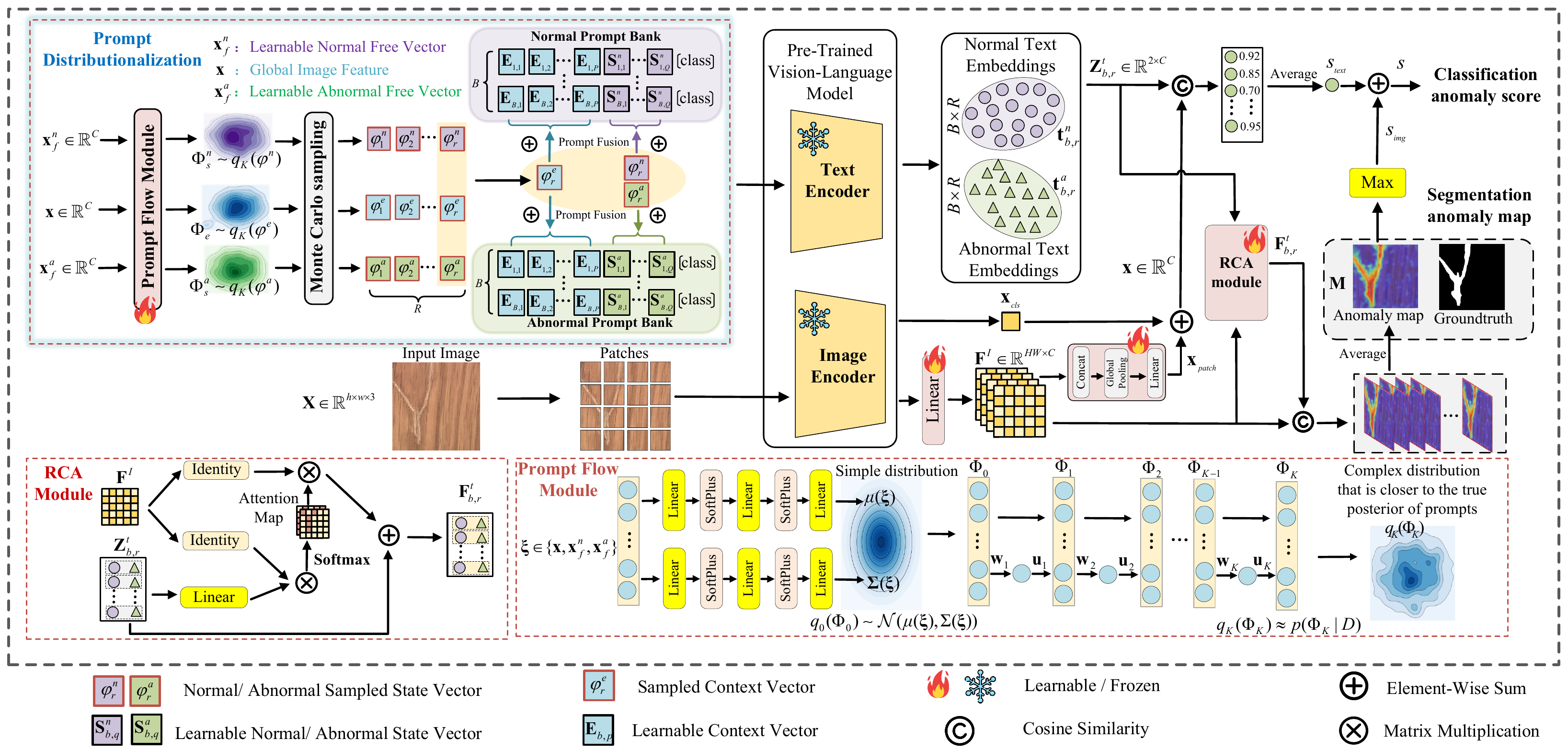}
	\caption{The framework of Bayes-PFL. First, a prompt bank with $B$ sets of learnable prompts is constructed (Section 3.2). Next, the prompt flow module converts the learnable prompts into a probabilistic distribution and generates $B \times R$ text embeddings through $R$ rounds of sampling (Sections 3.3 and 3.4), which are then aligned with image and patch features (Section 3.5).}
	\label{Fig2}
\end{figure*}
\subsection{Zero-shot Anomaly Detection}
 The earliest work, WinCLIP \cite{WinCLIP}, proposes a window-based strategy to aggregate classification results obtained from aligning text with sub-images at multiple scales. APRIL-GAN \cite{VAND} and CLIP-AD \cite{CLIPAD} use trainable adapter layers to map fine-grained patch features into a joint embedding space. The latter further employs feature surgery to mitigate opposite predictions. Additionally, several existing studies have integrated multiple foundation models, such as SAM \cite{SAM} and GroundingDINO \cite{Grounding}, to collectively enhance ZSAD performance, as demonstrated in SAA/SAA+ \cite{SAA} and ClipSAM \cite{ClipSAM}. Recently, an increasing number of prompt optimization-based methods have been proposed. AnomalyCLIP \cite{AnomalyCLIP} and Filo \cite{Filo} insert learnable vectors into the input text or CLIP encoder layers to avoid extensive engineering on hand-crafted prompt design. AdaCLIP \cite{AdaCLIP} and VCP-CLIP \cite{VCP} further utilize textual and visual hybrid prompts to enhance CLIP's anomaly perception capability. Different from these methods, our Bayes-PFL learns the distribution of textual prompts and leverages sampling to cover the prompt space to improve the ZSAD performance. 

\subsection{Prompt Design}
The design of textual prompts plays a pivotal role in applying VLMs to downstream tasks. Manually crafted textual prompts \cite{CLIP, CALIP, GroupViT,  Lseg,qiaoliang} and data-driven approaches to obtain learnable prompts \cite{CoOp,CoCoOp,DenseCLIP,KgCoOp, TaIDPT} in the natural scenario have both demonstrated effectiveness in zero-shot learning.  In the ZSAD task,  Compositional Prompt Ensemble \cite{WinCLIP, VAND, CLIPAD} and Prompt Optimization \cite{AnomalyCLIP, VCP, AdaCLIP} are two mainstream methods for text prompt design.
\par 
\textbf{Compositional Prompt Ensemble} generates diverse text prompts through the combination of state words and textual templates. In the representative work APRIL-GAN \cite{VAND}, approximately 35 templates and 5 state words are designed for abnormal descriptions, resulting in a total of $35\times 5$ text prompts. The text embeddings from these prompts are averaged to produce the final result, and a similar approach is applied to normal descriptions. 
\par  
\textbf{Prompt Optimization} replaces the context words in text prompts with learnable vectors or directly inserts these vectors into the encoder layers. As depicted in Figure \ref{Fig1}(a), the representative method AnomalyCLIP \cite{AnomalyCLIP} designs text prompts in the form of $[\mathbf{V}_1][\mathbf{V}_2]\cdots[\mathbf{V}_z][\text{state}][\text{object}]$, where $\mathbf{V}_i, i=1,2,\cdots, z$ are learnable vectors. However, the single-form prompt design and the lack of effective constraints on the prompt space limit its generalization capability to unseen categories.

%% file: sec/3_method.tex
\section{Method}
 Our approach follows the recent ZSAD task setting, where the model is trained in a supervised manner on seen categories $C^s$ from an industrial dataset and then directly tested on unseen categories $C^u$ from other industrial or medical datasets. Notably, $C^u \cap C^s = \emptyset$, and the object categories used for auxiliary training and testing come from different datasets with large domain gaps. 
\subsection{Overview}
Figure \ref{Fig2} illustrates the main framework of Bayes-PFL, which consists of three key designs: 1) Two-class prompt banks for normal and abnormal textual descriptions, 2) A prompt flow module for distribution learning, and 3) A RCA module for feature alignment. Given an input image $\mathbf{X}\in\mathbb{R}^{h\times w\times 3}$, the model first maps the patch-level image features to the joint embedding space through a single linear layer, producing $\mathbf{F}^I \in \mathbb{R}^{HW \times C}$, where $H = h / patchsize, W = w / patchsize$ and $C$ represents the dimensionality of the joint embedding space. Subsequently, the prompt flow module learns the image-specific distribution $q_K(\boldsymbol{\varphi}^e)$ for context words, and the image-agnostic distributions $q_K(\boldsymbol{\varphi}^n), q_K(\boldsymbol{\varphi}^a)$ for normal and abnormal state words. Monte Carlo sampling \cite{Sampling} is applied to these distributions, and the sampled results $\boldsymbol{\varphi}^e_r, \boldsymbol{\varphi}^n_r, \boldsymbol{\varphi}^a_r$ are fused with the prompt banks to generate diverse textual prompts. The text embeddings $\mathbf{Z}_{b,r}^t$ are aligned with $\mathbf{F}^I$ to obtain multiple anomaly segmentation results, which are averaged to derive the final anomaly map $\mathbf{M}$. Note that when aligning with $\mathbf{F}^I$, we additionally employ the proposed RCA module to refine the text embeddings to $\mathbf{F}_{b,r}^t$. 
\subsection{Two-class Prompt Banks}
We decompose the text prompt into three components: context words, state words, and class words. Unlike previous prompt optimization strategies \cite{AnomalyCLIP}, which manually design state words and only optimize context words, we argue that state words (e.g., good / damaged) still reside within an optimizable prompt space. Thus, in the word embedding space, the manually designed context and state words are replaced with learnable vectors. Inspired by Compositional Prompt Ensemble, normal and abnormal prompt banks are further constructed to expand the prompt space, with each containing $B$ distinct trainable prompts: 
\begin{gather}
	g_{b}^{n} = [\mathbf{E}_{b,1}][\mathbf{E}_{b,2}]\cdots[\mathbf{E}_{b,P}][\mathbf{S}_{b,1}^n]\cdots[\mathbf{S}_{b,Q}^n][\text{class}]  \\
	g_{b}^{a} = [\mathbf{E}_{b,1}][\mathbf{E}_{b,2}]\cdots[\mathbf{E}_{b,P}][\mathbf{S}_{b,1}^a]\cdots[\mathbf{S}_{b,Q}^a][\text{class}]
\end{gather}
where $b=1,2,\cdots,B$.  $\mathbf{E}_{b,i}\in \mathbb{R}^C,i=1,2,\cdots,P$ are learnable vectors designed to encode contextual information in the text. $\mathbf{S}_{b,j}^n, \mathbf{S}_{b,j}^a \in \mathbb{R}^C, j=1,2,\cdots,Q$ are learnable normal and abnormal state vectors, respectively. Note that we enforce orthogonality among the text embeddings generated from the same prompt bank, promoting diversity in the normal or anomalous semantics captured by different prompts. This constraint is implemented through the orthogonal loss $\mathcal{L}_{ort}$ introduced in Section 3.6.

\subsection{Prompt Distributionalization} 
Prompt distributionalization expresses a process of modeling and generating rich textual prompts by utilizing the prompt flow module, and fusing the sampled results from the distribution with the prompt banks. This process not only enhances the model's generalization to unseen categories by constraining the prompt space to ISD and IAD (distribution regularization), but also comprehensively covers the prompt space through distribution sampling.
\par 
\textbf{Bayesian inference.} Assume the auxiliary training dataset is $D = \{\mathbf{X}, Y_c, \mathbf{Y}_s\}$, where $\mathbf{X}\in\mathbb{R}^{h\times w\times 3}$ represents the input image, and $Y_c \in \{0,1\}$, $\mathbf{Y}_s \in \mathbb{R}^{h\times w}$ denote the image-level labels and pixel-level ground truth, respectively. To construct the text prompt distribution, the context, normal state, and abnormal state word embeddings are represented as $C$-dimensional random vectors $\Phi_e$, $\Phi_s^n$ and $\Phi_s^a$, respectively. For convenience, we denote $\Phi = \{\Phi_e,\Phi_s^n,\Phi_s^a \}$. The posterior probabilities are then computed as:
	\begin{equation}
	\label{eq16}
	p( \Phi | D ) = \frac{p(D | \Phi) p(\Phi)}{p(D)}
\end{equation}
Since computing the marginal likelihood $p(D)$ is intractable, variational inference \cite{variational} with distribution $q_\gamma(\Phi)$, parameterized by $\gamma$, is employed to approximate the posterior probability. Note that our method operates under the mean-field assumption, where all variables are mutually independent:
	$q_\gamma(\Phi) = q_\gamma(\boldsymbol{\varphi}^e)q_\gamma(\boldsymbol{\varphi}^n)q_\gamma(\boldsymbol{\varphi}^a)$. By applying Jensen's inequality, an upper bound is drived for the log marginal likelihood of the training data:
\begin{align}
	\log{p(D)} &= \log\int p(D | \Phi)p(\Phi)d\Phi\\
	&\hspace{-4em} \ge E_{q_\gamma(\Phi | D)}[\log p(D, \Phi) - \log q_\gamma(\Phi | D)] = -\mathcal{L}_e(D)
\end{align}
Thus, the variational distribution $q_\gamma(\Phi)$ can be obtained by minimizing the evidence lower bound (ELBO) loss $\mathcal{L}_e(D)$.
\par 
\textbf{Prompt flow module.} Richer posterior approximations enable us to better estimate the prompt distribution \cite{VINF}. Hence, a prompt flow module is designed to transform a simple probability distribution $q_0(\Phi_0)$ into a more complex one $q_K(\Phi_K)$ through a series of invertible mappings $h_k, k=1,2,\cdots K$. Specifically, this transformation follows $\Phi_K =  h_K(h_{K-1}(\cdots h_1(\Phi_0)))$, as illustrated in the bottom right corner of Figure \ref{Fig2}. For the model efficiency, we use a linear-time transformation $h(\Phi) = \Phi + \mathbf{u}g(\mathbf{w}^\text{T}\Phi + \mathbf{b})$, where $\mathbf{w},\mathbf{u} \in \mathbb{R}^C$, $\mathbf{b}\in \mathbb{R}$ are trainable parameters and $g(\cdot)$ is the \textit{Tanh} activation function \cite{Tanh}. After $K$ transformations, the final distribution is given by:
	\begin{equation}
	\label{eq16}
    \log q_K(\Phi_K) = \log q_0(\Phi_0) - \sum_{k=1}^{K}\log|1 + \mathbf{u}_k^\text{T}\phi(\Phi_k)|
\end{equation}
 where $\phi(\Phi) = g'(\mathbf{w}^T\Phi +\mathbf{b})\mathbf{w}$. By replacing $q_\gamma(\Phi | D)$ in Equation (5) with $q_K(\Phi_K)$, the objective function for optimizing the prompt flow module is computed as:
\begin{align}
	\mathcal{L}_p(D) &= E_{q_\gamma(\Phi | D)}[\log q_\gamma(\Phi | D) - \log p(D, \Phi)] \notag \\
	&\hspace{-2em}= E_{q_0(\Phi_0)}\left[\log q_0(\Phi_0) - \sum_{k=1}^{K}\log|1 + \mathbf{u}_k^\text{T}\phi(\Phi_k)|\right] \notag \\
	&\hspace{-2em}  - E_{q_0(\Phi_0)}[\log p(\Phi_K)] - E_{q_0(\Phi_0)}[\log p(D | \Phi_K)]
\end{align}
where the prior distribution follows $p = \mathcal{N}(0,\mathbf{I})$. The initial density $q_0$ is defined as a standard normal distribution: $q_0 = \mathcal{N}(\boldsymbol{\mu}(\boldsymbol{\xi}), \boldsymbol{\Sigma}(\boldsymbol{\xi}))$, where $\boldsymbol{\mu} \in \mathbb{R}^C$ and $diag(\boldsymbol{\Sigma}) =\boldsymbol{\sigma} \in \mathbb{R}^C$ are conditioned on the vector $\boldsymbol{\xi} \in \mathbb{R}^C$ and parameterized by three linear layers, with a Softplus activation function \cite{Tanh} applied between consecutive layers. 
\par
Depending on the different inputs $\boldsymbol{\xi}$ to the prompt flow module, two types of distributions are obtained: an \textbf{image-specific distribution} (ISD) and an \textbf{image-agnostic distribution} (IAD). For ISD, $\boldsymbol{\xi}$ is set as global image feature $\mathbf{x}$ to acquire a dynamic distribution, which models the context distribution and enhances generalization in unseen domains. For IAD, we set $\boldsymbol{\xi}$ as the learnable free vectors $\mathbf{x}_f^n$ and $\mathbf{x}_f^a$, respectively, while sharing the same network weights. This static distribution is used to learn a unified state semantics for both normal and abnormal conditions.
\par   
\textbf{Prompt sampling and fusion.} Monte Carlo sampling is used to sample $R$ iterations from the initial densities $q_0^e, q_0^n, q_0^a$. The sampled results are processed through the prompt flow module to obtain $ \boldsymbol{\varphi}^e_{r},  \boldsymbol{\varphi}^n_{r},  \boldsymbol{\varphi}^a_{r}, r = 1,2,\cdots, R$, which represent sampled context vector, normal sampled state vector and abnormal sampled state vector, respectively. The fusion process is formulated as:
\begin{equation}
\begin{aligned}
	g_{b,r}^{n} = [\mathbf{E}_{b,1} + \boldsymbol{\varphi}^e_{r}][\mathbf{E}_{b,2} + \boldsymbol{\varphi}^e_{r}]\cdots[\mathbf{E}_{b,P}+ \boldsymbol{\varphi}^e_{r}] \\
	[\mathbf{S}_{b,1}^n + \boldsymbol{\varphi}^n_{r}]\cdots[\mathbf{S}_{b,Q}^n + \boldsymbol{\varphi}^n_{r}][\text{class}]	
\end{aligned}
\end{equation}
\begin{equation}
	\begin{aligned}
		g_{b,r}^{a} = [\mathbf{E}_{b,1} + \boldsymbol{\varphi}^e_{r}][\mathbf{E}_{b,2} + \boldsymbol{\varphi}^e_{r}]\cdots[\mathbf{E}_{b,P}+ \boldsymbol{\varphi}^e_{r}] \\
		[\mathbf{S}_{b,1}^a + \boldsymbol{\varphi}^a_{r}]\cdots[\mathbf{S}_{b,Q}^a + \boldsymbol{\varphi}^a_{r}][\text{class}]	
	\end{aligned}
\end{equation}
where $g_{b,r}^{n},\ g_{b,r}^{a}$ represent the text prompt obtained from the $r\text{-}th$ sampling of the $b\text{-}th$ prompt in the prompt banks, respectively.  After sampling and fusion, the number of prompts in each bank increases to $B\times R$ at a very low cost. Note that to ensure proper gradient backpropagation after discrete sampling, the reparameterization method \cite{VAE} is employed during the optimization process. 
\subsection{Residual Cross-modal Attention Module}
The normal and abnormal text embeddings, derived from the text prompts in Equations (8) and (9) through the text encoder, are represented as $\mathbf{t}_{b,r}^n$ and $\mathbf{t}_{b,r}^a$, respectively. Due to the design of image-specific distribution and Monte Carlo sampling, the generated text embeddings are dynamic during both training and inference. This presents challenges in aligning text embeddings $\mathbf{Z}_{b,r}^t \in \mathbb{R}^{2\times C}$ with fine-grained patch embeddings $\mathbf{F}^I\in \mathbb{R}^{HW\times C}$ in anomaly segmentation task, where $\mathbf{Z}^t_{b,r}$ is the concatenation of $\mathbf{t}_{b,r}^n$ and $\mathbf{t}_{b,r}^a$ along the sequence dimension. Therefore, the RCA module is designed to promote cross-modal interaction between textual and fine-grained image features, which can be formulated as follows:
\begin{equation}
	\begin{aligned}
	  \mathbf{F}_{b,r}^t = \mathbf{Z}^t_{b,r} + softmax (\mathbf{Q}_{b,r}(\mathbf{F}^I)^\text{T} / \sqrt{C}) \mathbf{F}^I
	\end{aligned}
\end{equation}
where $\mathbf{F}_{b,r}^t \in \mathbb{R}^{2\times C}$ is the refined text embeddings. The query embedding $\mathbf{Q}_{b,r} = \mathbf{Z}^t_{b,r}\mathbf{W}$, where $\mathbf{W}\in \mathbb{R}^{C\times C}$ is the weight matrix of a linear mapping layer. The RCA module introduces residuals to: 1) fuse original textual features with cross-modal features for better zero-shot performance and 2) facilitate gradient backpropagation to the prompt bank, improving the optimization of learnable prompts.
\subsection{Anomaly Map and Anomaly Score}
\textbf{Pixel-level anomaly map.}  We extract $L$-layer patch-level features $\mathbf{F}^I_l$ and align them with the refined text embeddings $\mathbf{F}_{l,b,r}^{t}, l=1,2,\cdots, L$. The anomaly map from the $l\text{-}th$ layer is computed as:
\begin{equation}
	\mathbf{M}_{l,b,r} = softmax (Up(\widetilde{\mathbf{F}}^I_l \widetilde{\mathbf{F}}_{l,b,r}^{t\text{T}}))
\end{equation}
where $\widetilde{(\cdot)}$ denotes the $L_2$-normalization operation along the embedding dimension and $Up(\cdot)$ is the unsampling operation. The final result $\mathbf{M}\in \mathbb{R}^{h\times w}$ is obtained by averaging the anomaly maps derived from aligning the $L$-layer patch features with $B\times R$ refined text embeddings:
 \begin{equation}
	\mathbf{M} = \frac{1}{LBR}\sum_{l=1}^{L}\sum_{b=1}^{B}\sum_{r = 1}^{R} \mathbf{M}_{l,b,r}
\end{equation}
\par 
\textbf{Image-level anomaly score.} For anomaly classification, the anomaly values come from both the text and image branches. The anomaly score in the text branch is computed as: $s_{text} = (\frac{1}{BR})\sum_{b}^{B}\sum_{r}^{R} softmax (\widetilde{\mathbf{x}}\widetilde{\mathbf{Z}}_{b,r}^{t\text{T}})$, where $\mathbf{x} = \mathbf{x}_{{cls}} + \mathbf{x}_{{patch}}$ is the global image embedding. Note that $\mathbf{x}_{{cls}}$ refers to the global image features obtained from the class token of the vanilla image encoder, while $\mathbf{x}_{{patch}}$ comes from the fusion of fine-grained patch features. Specifically, the patch features $\mathbf{F}^I_l \in \mathbb{R}^{H\times W \times C}, l=1,2,\cdots, L$ from different layers are first concatenated along the channel dimension. Then, global average pooling is applied along the spatial dimensions, followed by a linear layer that maps the results to $\mathbf{x}_{patch}$. The anomaly score from the image branch is derived from the maximum value of the anomaly map: $s_{img} =\max(\mathbf{M})$. The final image-level anomaly score is then expressed as: $s = s_{text} + s_{img}$.

\subsection{Loss Function}

During the training stage, a single Monte Carlo sampling is performed to enhance efficiency. Let the text prompts from $B$ samples be denoted as $[g_{b,1}^{n}, g_{b,1}^{a}], b\in\{1,2,\cdots B\}$, and their corresponding text embeddings as $[\mathbf{t}_{b,1}^{n}, \mathbf{t}_{b,1}^{a}], b\in\{1,2,\cdots B\}$. To enhance the diversity of learnable prompts in prompt banks, an orthogonal loss is designed on the text embeddings as follows:
 \begin{equation}
	\mathcal{L}_{ort} = \sum_{i=1}^{B}\sum_{j=1, j\ne i}^{B} \left\{(\langle \mathbf{t}_{i,1}^n,\mathbf{t}_{j,1}^n \rangle)^2 + (\langle \mathbf{t}_{i,1}^a,\mathbf{t}_{j,1}^a \rangle)^2\right\}
\end{equation}
in which $\langle \cdot,\cdot \rangle$ denotes the cosine similarity. The final loss function can be expressed as: 
 \begin{equation}
\mathcal{L} = \mathcal{L}_{ort} + \mathcal{L}_p
\end{equation}
where the prompt flow loss $\mathcal{L}_p$ is computed using Equation (7). The first two terms of $\mathcal{L}_p$ are distribution regularization terms that refine the initial distribution $q_0(\Phi_0)$ into a simple prior $p(\Phi_K)$, effectively controlling the distribution of latent variables. This regularization ensures that the prompt flow module can learn an effective prompt distribution, mitigating overfitting while enhancing the generalization of the prompts. The third term, $E_{q_0(\Phi_0)}[\log p(D | \Phi_K)]$, maximizes the log-likelihood of the data, which is approximated by the sum of classification loss (for text-image alignment) and segmentation loss (for text-patch alignment). Specifically, our Bayes-PFL employs a cross-entropy loss for classification, along with the sum of Focal loss \cite{Focal} and Dice loss \cite{Dice} for segmentation. More details about the loss function can be found in Appendix A.1.

%% file: sec/4_experiments.tex
\section{Experiments}
\subsection{Experimental Setup}
% \usepackage{multirow}
\begin{table*}[t]
	\caption{Comparison with existing state-of-the-art methods. The best results are marked in \textcolor{darkred}{red}, while the second-best are indicated in \textcolor{blue}{blue}. }
	\centering
	\label{Tab1}
	\renewcommand{\arraystretch}{1.0}
	\resizebox{1.9\columnwidth}{!}
	{
		\begin{tabular}{>{\centering\arraybackslash}p{1.2cm}>{\centering\arraybackslash}p{2.3cm}>{\centering\arraybackslash}p{2.3cm} *{3}{>{\centering\arraybackslash}p{2.6cm}}>{\centering\arraybackslash}p{2.8cm}>{\centering\arraybackslash}p{2.6cm}>{\columncolor{lightgreen}\centering\arraybackslash}p{2.6cm}}
			\toprule
			Domain	&   Metric                                           &    Dataset           & WinCLIP \cite{WinCLIP}            & APRIL-GAN \cite{VAND}          & CLIP-AD \cite{CLIPAD}           & AnomalyCLIP \cite{AnomalyCLIP}        & AdaCLIP \cite{AdaCLIP}             & \textbf{Bayes-PFL}               \\   \midrule
			\multirow{14}{*}{Industrial} & \multirow{7}{*}{\makecell[c]{Image-level \\ (AUROC, F1-Max, \\ AP)}} & MVTec-AD         & (91.8, \textcolor{blue}{92.9},   95.1) & (86.1, 90.4, 93.5)   & (89.8, 91.1, 95.3)   & (91.5, 92.8, 96.2)   & (\textcolor{blue}{92.0}, 92.7, \textcolor{blue}{96.4})  & (\textcolor{darkred}{92.3}, \textcolor{darkred}{93.1}, \textcolor{darkred}{96.7}) \\
			&                                              & VisA          & (78.1, 79.0, 77.5)   & (78.0, 78.7, 81.4)   & (79.8, 79.2, 84.3)   & (82.1, 80.4, \textcolor{blue}{85.4})   & (\textcolor{blue}{83.0}, \textcolor{blue}{81.6}, 84.9)  & (\textcolor{darkred}{87.0}, \textcolor{darkred}{84.1}, \textcolor{darkred}{89.2}) \\
			&                                              & BTAD          & (83.3, 81.0, 84.1)   & (74.2, 70.0, 71.7)   & (85.8, 81.7, 85.2)   & (89.1, 86.0, 91.1)   & (\textcolor{blue}{91.6}, \textcolor{blue}{88.9}, \textcolor{blue}{92.4})  & (\textcolor{darkred}{93.2}, \textcolor{darkred}{91.9}, \textcolor{darkred}{96.5}) \\
			&                                              & KSDD2         & (93.5, 71.4, 77.9)   & (90.3, 70.0, 74.4)   & (95.2, 84.4, 88.2)   & (92.1, 71.0, 77.8)  & (\textcolor{blue}{95.9}, \textcolor{blue}{84.5}, \textcolor{blue}{95.9})  & (\textcolor{darkred}{97.3}, \textcolor{darkred}{92.3}, \textcolor{darkred}{97.9}) \\
			&                                              & RSDD          & (85.3, 73.5, 65.3)   & (73.1, 59.7, 50.5)   & (88.3, 74.1, \textcolor{blue}{73.9})   & (73.5, 59.0, 55.0)   & (\textcolor{blue}{89.1}, \textcolor{blue}{75.0}, 70.8)  & (\textcolor{darkred}{94.1}, \textcolor{darkred}{89.6}, \textcolor{darkred}{92.3}) \\
			&                                              & DAGM          & (89.6, 86.4, 90.4)   & (90.4, 86.9, 90.1)   & (90.8, 88.4, 90.5)   & (95.6, 93.2, 94.6)   & (\textcolor{blue}{96.5}, \textcolor{blue}{94.1}, \textcolor{blue}{95.7})  & (\textcolor{darkred}{97.7}, \textcolor{darkred}{95.7}, \textcolor{darkred}{97.0}) \\
			&                                              & DTD-Synthetic & (\textcolor{blue}{95.0}, 94.3,  \textcolor{blue}{97.9}) & (83.9, 89.4, 93.6)   & (91.5, 91.8, 96.8)   & (94.5, \textcolor{blue}{94.5}, 97.7)   & (92.8, 92.2, 97.0)  & (\textcolor{darkred}{95.1}, \textcolor{darkred}{95.1}, \textcolor{darkred}{98.4}) \\ \cmidrule{2-9}
			& \multirow{7}{*}{\makecell[c]{Pixel-level \\ (AUROC, PRO, \\ AP)}}  & MVTec-AD         & (85.1, 64.6, 18.0)   & (87.6, 44.0, \textcolor{blue}{40.8})  & (89.8, 70.6, 40.0)   & (\textcolor{blue}{91.1}, \textcolor{blue}{81.4}, 34.5)   & (86.8, 33.8, 38.1)  & (\textcolor{darkred}{91.8}, \textcolor{darkred}{87.4}, \textcolor{darkred}{48.3})  \\
			&                                              & VisA          & (79.6, 56.8,  5.0)   & (94.2, 86.8, 25.7)   & (95.0, 86.9, 26.3)   & (\textcolor{blue}{95.5}, \textcolor{blue}{87.0}, 21.3)   & (95.1, 71.3, \textcolor{blue}{29.2})  & (\textcolor{darkred}{95.6}, \textcolor{darkred}{88.9}, \textcolor{darkred}{29.8}) \\
			&                                              & BTAD          & (71.4, 32.8, 11.2)   & (91.3, 23.0, 32.9)   & (93.1, 59.8, \textcolor{blue}{46.7})   & (\textcolor{blue}{93.3}, \textcolor{blue}{69.3}, 42.0)   & (87.7, 17.1, 36.6)  & (\textcolor{darkred}{93.9}, \textcolor{darkred}{76.6}, \textcolor{darkred}{47.1}) \\
			&                                              & KSDD2         & (97.9, 91.2, 17.1)   & (97.9, 51.1, \textcolor{blue}{61.6})   & (99.3, 85.4, 58.9)   & (\textcolor{blue}{99.4}, \textcolor{blue}{92.7}, 41.8)   & (96.1, 70.8, 56.4)  & (\textcolor{darkred}{99.6}, \textcolor{darkred}{97.6}, \textcolor{darkred}{73.7}) \\
			&                                              & RSDD          & (95.1, 75.4, 2.1)    & (99.4, 64.9, 30.6)   & (99.2, 90.1, 31.9)   & (99.1, \textcolor{blue}{92.0}, 19.1) & (\textcolor{blue}{99.5}, 50.5, \textcolor{blue}{38.2})  & (\textcolor{darkred}{99.6}, \textcolor{darkred}{98.0}, \textcolor{darkred}{39.1}) \\
			&                                              & DAGM          & (83.2, 55.4, 3.1)    & (\textcolor{blue}{99.2}, 44.7, 42.6)   & (99.0, 83.1, 40.7)   & (99.1, \textcolor{blue}{93.6}, 29.5)   & (97.0, 40.9, \textcolor{darkred}{44.2})  & (\textcolor{darkred}{99.3}, \textcolor{darkred}{98.0}, \textcolor{blue}{43.1}) \\
			&                                              & DTD-Synthetic & (82.5, 55.4, 11.6)   & (96.6, 41.6, \textcolor{blue}{67.3})   & (97.1, 68.7, 62.3)   & (\textcolor{blue}{97.6}, \textcolor{blue}{88.3}, 52.4)   & (94.1, 24.9, 52.8)  & (\textcolor{darkred}{97.8}, \textcolor{darkred}{94.3}, \textcolor{darkred}{69.9}) \\ \midrule
			\multirow{8}{*}{Medical}   & \multirow{3}{*}{\makecell[c]{Image-level \\ (AUROC, F1-Max, \\ AP)}} & HeadCT        & (83.7, 78.8, 81.6)   & (89.3, 82.0, 89.6)   & (93.8, \textcolor{blue}{90.5}, 92.2) & (\textcolor{blue}{95.3}, 89.7, \textcolor{blue}{95.2})   & (93.4, 86.5, 92.2)  & (\textcolor{darkred}{96.5}, \textcolor{darkred}{92.9}, \textcolor{darkred}{95.5}) \\
			&                                              & BrainMRI      & (92.0, 84.2, 90.7)   & (89.6, 85.3, 84.5)   & (92.8, 88.7, 85.5)   & (\textcolor{blue}{96.1}, \textcolor{blue}{92.3}, 92.3)   & (94.9, 90.4, \textcolor{darkred}{94.2})  & (\textcolor{darkred}{96.2}, \textcolor{darkred}{92.8}, \textcolor{blue}{92.4}) \\
			&                                              & Br35H         & (80.5, 74.1, 82.2)   & (93.1, 85.4, 92.9)   & (96.0, 90.8, 95.5)   & (\textcolor{blue}{97.3}, \textcolor{blue}{92.4}, \textcolor{blue}{96.1})   & (95.7, 89.1, 95.7 ) & (\textcolor{darkred}{97.8}, \textcolor{darkred}{93.6}, \textcolor{darkred}{96.2}) \\ \cmidrule{2-9}
			&  \multirow{5}{*}{\makecell[c]{Pixel-level \\ (AUROC, PRO, \\ AP)}}  & ISIC          & (83.3, 55.1, 62.4)   & (85.8, 13.7, 69.8)   & (81.6, 29.0, 65.5)   & (\textcolor{blue}{88.4}, \textcolor{blue}{78.1}, \textcolor{blue}{74.4})   & (85.4, 5.3, 70.6)   & (\textcolor{darkred}{92.2}, \textcolor{darkred}{87.6}, \textcolor{darkred}{84.6}) \\
			&                                              & CVC-ColonDB   & (64.8, 28.4, 14.3)   & (78.4, 28.0, 23.2)   & (80.3, 58.8, 23.7)   & (\textcolor{blue}{81.9}, \textcolor{blue}{71.4}, \textcolor{blue}{31.3})   & (79.3, 6.5, 26.2)   & (\textcolor{darkred}{82.1}, \textcolor{darkred}{76.1}, \textcolor{darkred}{31.9}) \\
			&                                              & CVC-ClinicDB  & (70.7, 32.5, 19.4)   & (\textcolor{blue}{86.0}, 41.2,   38.8) & (85.8, \textcolor{blue}{69.7}, 39.0) & (85.9, 69.6, \textcolor{blue}{42.2})   & (84.3, 14.6, 36.0)  & (\textcolor{darkred}{89.6}, \textcolor{darkred}{78.4}, \textcolor{darkred}{53.2}) \\
			&                                              & Endo          & (68.2, 28.3, 23.8)   & (84.1, 32.3, 47.9)   & (85.6, 57.0, \textcolor{blue}{51.7})   & (\textcolor{blue}{86.3}, \textcolor{blue}{67.3}, 50.4)   & (84.0, 10.5, 44.8)  & (\textcolor{darkred}{89.2}, \textcolor{darkred}{74.8}, \textcolor{darkred}{58.6}) \\
			&                                              & Kvasir        & (69.8, 31.0, 27.5)   & (80.2, 27.1, 42.4)   & (\textcolor{blue}{82.5}, 48.1, \textcolor{blue}{46.2})   & (81.8, \textcolor{blue}{53.8},   42.5) & (79.4, 12.3, 43.8) & (\textcolor{darkred}{85.4}, \textcolor{darkred}{63.9}, \textcolor{darkred}{54.2})
			 \\ \bottomrule
		\end{tabular}
	}
\end{table*}
\textbf{Datasets.} To evaluate the ZSAD performance of the model, experiments on 15 real-world datasets from the industrial and medical domains are conducted. Specifically, we utilize seven commonly used industrial anomaly detection datasets, including MVTec-AD \cite{MVTec}, VisA \cite{VisA}, BTAD \cite{BTAD}, KSDD2 \cite{KSDD2}, RSDD \cite{RSDD}, DAGM \cite{DAGM} and DTD-Synthetic \cite{DTD}. Eight different datasets in the medical domain are used, including HeadCT \cite{HeadCT}, BrainMRI \cite{BrainMRI}, and Br35H \cite{Br35H} for brain tumor classification, CVC-ColonDB \cite{ColonDB}, CVC-ClinicDB \cite{ClinicDB}, Endo \cite{Endo}, and Kvasir \cite{kvasir} for colon polyp segmentation and ISIC \cite{ISIC} for skin cancer detection. Following \cite{CLIPAD, VCP}, we perform auxiliary training on one industrial dataset and directly infer on other industrial and medical datasets. Since the categories in VisA do not overlap with those in the other datasets, we use VisA as the auxiliary training set. To assess VisA itself, we fine-tune our model on the MVTec-AD dataset. More details about the datasets can be found in Appendix C.
\par 
\textbf{Evaluation Metrics.} For classification, we report image-level metrics: Area Under the Receiver Operating Characteristic (AUROC), the maximum F1 score at the optimal threshold (F1-Max) and the average precision (AP). For segmentation, pixel-level AUROC, AP and Per-Region Overlap (PRO) are employed to evaluate the model's ability. 
\par 
\textbf{Implementation Details.} The publicly available CLIP (\textit{ViT-L-14-336}) pretrained by OpenAI \cite{CLIP} is adopted in this study. Following previous works \cite{VAND, AnomalyCLIP, AdaCLIP}, we standardize the resolution of input images to $518 \times 518$ and select the 6th, 12th, 18th, 24th layers from the 24-layer image encoder to extract patch embeddings. The number of prompts $B$ in the prompt banks is set to 3 and flow length $K$ is set to 10 by default. The length of learnable context vectors $P$ and learnable state vectors $Q$ are both set to 5. During the auxiliary training and inference stages, the number of Monte Carlo sampling interations $R$ is set to 1 and 10, respectively. We use the Adam \cite{Adam} optimizer to train our Bayes-PFL for a total of 20 epochs, with an initial learning rate of 0.0001 and a batch size of 32. All experiments are conducted on a single NVIDIA 3090 GPU, with 5 random seeds selected. The experimental results are the averages of different object categories from the target dataset. More implementation details can be found in Appendix A.1.
\par 

\subsection{Comparison with State-of-the-art methods}
\begin{figure*}[t]
	\begin{minipage}[b]{1.16\columnwidth}  % Left column (first figure)
		\centering
		\includegraphics[width=1.01\columnwidth]{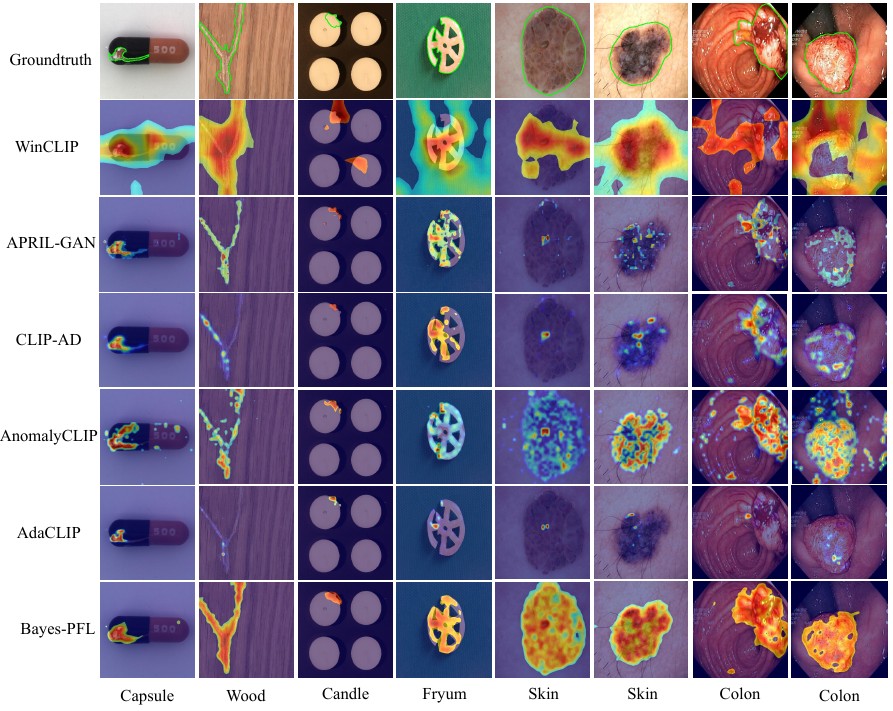}
		\caption{Qualitative comparison of anomaly segmentation results across different ZSAD methods. The images in the first four columns are from the industrial datasets MVTec-AD \cite{MVTec}, VisA \cite{VisA}, whereas those in the last four columns are from the medical datasets ISIC \cite{ISIC}, Kvasir \cite{kvasir}, and CVC-ClinicDB \cite{ClinicDB}.}
		\label{Fig3}
	\end{minipage}
	\hfill
	\begin{minipage}[b]{0.9\columnwidth}  % Right column (two figures stacked)
		\centering
		\includegraphics[width=0.80\columnwidth]{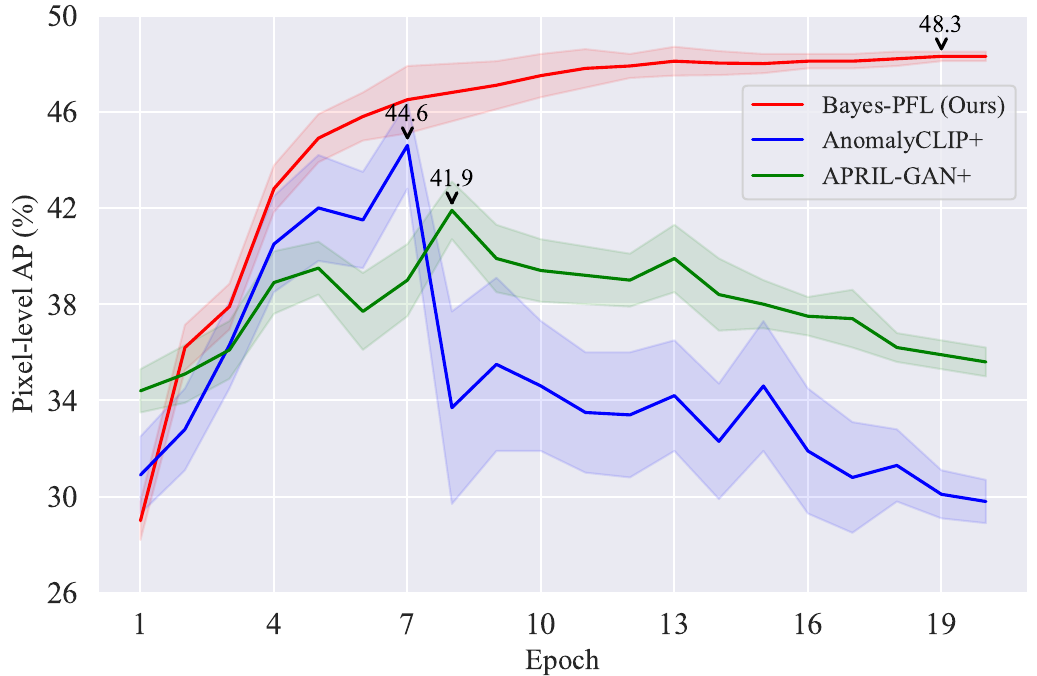}
		\caption{Pixel-level AP during the test stage at different epoches on the MVTec-AD dataset.}
		\label{Fig4}
		
		\vspace{0.1cm}  % Vertical space between the two figures
		
		\includegraphics[width=0.9\columnwidth]{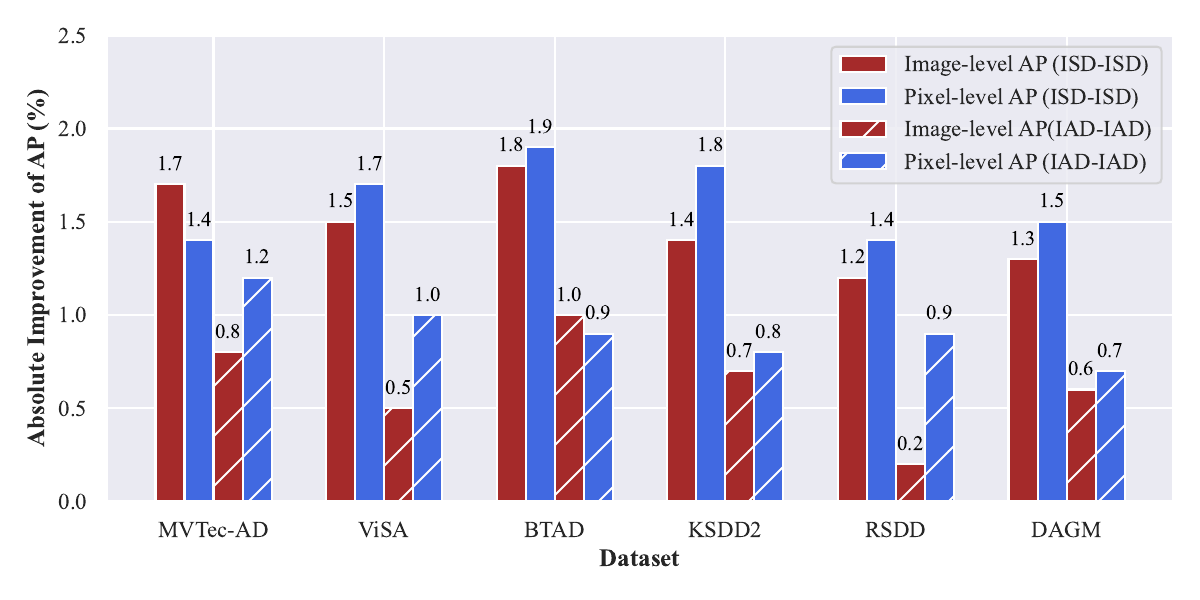}
		\caption{Absolute AP improvement of using our ISD-IAD compared to ISD-ISD and IAD-IAD.}
		\label{Fig5}
	\end{minipage}
\end{figure*}
In this study, five SOTA methods are employed for comparison with our Bayes-PFL, including  WinCLIP \cite{WinCLIP}, APRIL-GAN \cite{VAND}, CLIP-AD \cite{CLIPAD}, AnomalyCLIP \cite{AnomalyCLIP}, and AdaCLIP \cite{AdaCLIP}. To ensure a fair comparison, we use the same backbone, input image resolution, and experimental settings (training on VisA and testing on other datasets) for all methods. More details regarding the comparison of these methods are shown in Appendix A.2.
\par 
\textbf{Quantitative Comparison.} Table \ref{Tab1} presents the quantitative results of ZSAD across different datasets in both industrial and medical domains. Our Bayes-PFL achieves state-of-the-art performance on image-level and pixel-level metrics across nearly all datasets in comparison with other methods. For example, on the medical dataset ISIC, our Bayes-PFL outperforms the second-best approach by 3.8\%, 9.5\%, and 10.2\% in pixel-level AUROC, PRO, and AP, respectively. The poor PRO performance of AdaCLIP is attributed to the high variance in its predicted anomaly scores and its limited ability to effectively detect large anomalous regions. In contrast, Bayes-PFL explores a broader prompt space by sampling from the distribution, enhancing the detection of large anomalous regions. Note that the anomaly detection results on the medical dataset are directly derived from the fine-tuned weights of the industrial dataset VisA, indicating that the feature spaces for anomalies in these two domains overlap to some extent.
\par 
\textbf{Qualitative Comparison.} Figure \ref{Fig3} visualizes anomaly maps from both industrial and medical datasets. It can be observed that our Bayes-PFL achieves more accurate segmentation results and more complete anomaly region localization compared to other methods. Especially in categories from the medical domain (e.g., colon, skin), our method demonstrates a significant advantage. These results indicate that the learned prompt distribution can cover a more comprehensive prompt space and capture the anomaly semantics in images from different novel categories. 
\par 
\textbf{Performance comparison at different training epochs.} In Figure \ref{Fig4}, we compare APRIL-GAN \cite{VAND} and AnomalyCLIP \cite{AnomalyCLIP}, which employ two different text prompt design strategies, with our Bayes-PFL across various training epochs. The pixel-level AP during testing is reported to evaluate ZSAD performance. For a fair comparison, we modify these methods by replacing only their text prompt designs while keeping all other components identical to Bayes-PFL. These modified versions are referred to as APRIL-GAN+ and AnomalyCLIP+, respectively. Our Bayes-PFL achieves stable and consistently improving AP throughout training, surpassing the other two methods and highlighting the advantages of our prompt flow learning approach. In contrast, APRIL-GAN+ and AnomalyCLIP+ exhibit performance declines after reaching their peak AP values, with the latter showing a more significant degradation. In Appendix B.4, we further explore this phenomenon and analyze the reasons behind it.
\par 
\textbf{Remark.} Our Bayes-PFL outperforms other SOTA approaches for three main reasons: 1) The prompt flow module effectively regularizes the prompt space with ISD and IAD, thereby enhancing the model's generalization performance on unseen categories; 2) The distribution sampling adaptively generates diverse text prompts to cover the prompt space. The ensemble of results from rich prompts enables better ZSAD performance in Bayes-PFL; 3) The RCA module facilitates cross-modal intersection and enhances the alignment capability of different modalities. 
\par 
\subsection{Image-specific vs. Image-agnostic distribution}
In Bayes-PFL, ISD and IAD are modeled in the context and state word embedding spaces, respectively, referred to collectively as ISD-IAD. A natural question arises: why not use ISD-ISD or IAD-IAD instead? In Figure \ref{Fig5}, we compare the absolute AP improvement achieved by using ISD-IAD over ISD-ISD and IAD-IAD across six industrial datasets. Higher values indicate a more pronounced advantage of our approach. It can be observed that all metrics are positive, indicating that ISD-IAD outperforms the other two distribution design strategies. This is because the text prompts need to learn unified normal and abnormal semantics across different categories, making the static IAD more suitable for capturing category-invariant state representations. The dynamic ISD incorporates visual features, aiding in the acquisition of rich contextual information (e.g., object background) in the context word embeddings.
\subsection{Ablation}
In this subsection, ablation experiments on the MVTec-AD dataset are conducted to further investigate the impact of different settings on the proposed Bayes-PFL.
\par 
\begin{table}[t]
	\caption{Ablation on different components.}
	\centering
	\label{Tab2}
	\renewcommand{\arraystretch}{1}
	\resizebox{0.9\columnwidth}{!}
	{
		\begin{tabular}{>{\centering\arraybackslash}p{2.2cm}>{\centering\arraybackslash}p{2cm}*{4}{>{\centering\arraybackslash}p{1cm}}}
			\toprule
			&              & \multicolumn{2}{c}{Image-level} & \multicolumn{2}{c}{Pixel- level} \\  \cmidrule(lr){3-4} \cmidrule(lr){5-6}
			&              & AUROC           & AP            & AUROC           & AP             \\ \midrule
			\multirow{4}{*}{\makecell[c]{Module \\ Ablation}}           & w/o ISD       & 89.1            & 94.5          & 88.3            & 46.5           \\
			& w/o IAD       & 90.3            & 95.2          & 89.5            & 47.2           \\
			& w/o ISD, IAD   & 88.6            & 93.3          & 87.1            & 45.6           \\
			& w/o RCA      & 91.8            & 95.9          & 89.7            & 46.1           \\   \midrule
			Loss Ablation                              & w/o $\mathcal{L}_{ort}$   & 91.2            & 96.0          & 90.2            & 47.3           \\  \midrule
			\multirow{4}{*}{\makecell[c]{Classification \\ Ablation}} & w/o $\mathbf{x}_{cls}$   & 89.3            & 94.4          & ---             & ---            \\
			& w/o $\mathbf{x}_{patch}$ & 90.9            & 95.8          & ---             & ---            \\
			& w/o $s_{text}$  & 89.0            & 95.1          & ---             & ---            \\
			& w/o $s_{img}$   & 90.9            & 95.3          & ---             & ---            \\    \midrule \rowcolor{lightgreen}
			& Bayes-PFL    & \textbf{92.3}            & \textbf{96.7}         & \textbf{91.8}            & \textbf{48.3}            \\  \bottomrule
		\end{tabular}
	}
\end{table}

\begin{table}[t]
	\caption{Ablation on the number of Monte Carlo sampling iterations during inference. (mean$\pm$std)}
	\centering
	\label{Tab3}
	\renewcommand{\arraystretch}{1}
	\resizebox{0.9\columnwidth}{!}
	{
		\begin{tabular}{>{\centering\arraybackslash}p{0.3cm}*{4}{>{\centering\arraybackslash}p{1.3cm}}>{\centering\arraybackslash}p{1.2cm}}
			\toprule
			\multirow{2}{*}{$R$} & \multicolumn{2}{c}{Image-level} & \multicolumn{2}{c}{Pixel-level} & \multirow{2}{*}{\makecell[c]{Time \\ (ms)}} \\
			\cmidrule(lr){2-3}  \cmidrule(lr){4-5}
			& AUROC          & AP             & AUROC          & AP             &                       \\  \midrule
			1                  & 89.5{\footnotesize{$\pm$1.1}}       & 94.1{\footnotesize{$\pm$1.2}}       & 87.1{\footnotesize{$\pm$1.3}}       & 45.4{\footnotesize{$\pm$0.6}}      & 132.2{\footnotesize{$\pm$1.6}}                   \\
			3                  & 90.0{\footnotesize{$\pm$1.0}}      & 94.8{\footnotesize{$\pm$0.9}}       & 88.2{\footnotesize{$\pm$1.2}}      & 46.1{\footnotesize{$\pm$0.6}}      & 200.3{\footnotesize{$\pm$1.4}}                    \\
			5                  & 91.1{\footnotesize{$\pm$0.8}}       & 95.4{\footnotesize{$\pm$0.9}}     & 89.5{\footnotesize{$\pm$1.0}}       & 46.8{\footnotesize{$\pm$0.5}}       & 254.2{\footnotesize{$\pm$1.4}}                    \\
			7                  & 92.0{\footnotesize{$\pm$0.7}}      & 96.3{\footnotesize{$\pm$0.8}}      & 90.9{\footnotesize{$\pm$0.8}}       & 47.5{\footnotesize{$\pm$0.5}}      & 297.4{\footnotesize{$\pm$1.7}}                    \\   \rowcolor{lightgreen}
			10                 & 92.7{\footnotesize{$\pm$0.4}}     & 96.8{\footnotesize{$\pm$0.5}}      & 91.8{\footnotesize{$\pm$0.6}}      & 48.3{\footnotesize{$\pm$0.4}}      & 388.5{\footnotesize{$\pm$1.5}}                    \\
			13                 & 92.7{\footnotesize{$\pm$0.4}}      & 96.9{\footnotesize{$\pm$0.5}}       & 91.9{\footnotesize{$\pm$0.5}}      & 48.4{\footnotesize{$\pm$0.4}}      & 467.5{\footnotesize{$\pm$1.5}}                   \\
			16                 & 92.8{\footnotesize{$\pm$0.3}}      & 96.9{\footnotesize{$\pm$0.4}}      & 92.0{\footnotesize{$\pm$0.4}}       & 48.6{\footnotesize{$\pm$0.3}}     & 621.6{\footnotesize{$\pm$1.6}}                   \\
			20                 & 92.9{\footnotesize{$\pm$0.2}}      & 97.0{\footnotesize{$\pm$0.2}}       & 92.1{\footnotesize{$\pm$0.2}}      & 48.9{\footnotesize{$\pm$0.1}}      & 726.1{\footnotesize{$\pm$1.7}}              \\ \bottomrule    
		\end{tabular}
	}
\end{table}

\textbf{Influence of different components.} Ablation studies on modules, loss functions, and classifications are conducted separately, as summarized in Table \ref{Tab2}. The module ablation study shows that ZSAD performance declines when the RCA module or any distribution type in the prompt flow module is removed. Notably, omitting ISD results in a greater performance drop than omitting IAD in all metrics, indicating that context distribution modeling conditioned on the image has a greater impact on generalization ability. Not using the orthogonal loss $\mathcal{L}_{ort}$ also leads to a decrease of approximately 1\% in both image-level and pixel-level metrics. In the classification ablation study, the best performance is achieved when all components are included. This suggests that the results from the text and image branches complement each other, and integrating fine-grained patch features into the global image features $\mathbf{x}_{cls}$ from the vanilla CLIP model helps enhance classification performance.
\par 
\textbf{Influence of the number of sampling iterations.} The ZSAD performance on novel categories under different sampling iterations $R$ is presented in Table \ref{Tab3}. Additionally, the standard deviation and average inference time per image (calculated from 50 images) are reported to assess the stability of the results and inference efficiency. As the number of sampling iterations increases, our method demonstrates improved zero-shot generalization performance and stability, but at the cost of longer inference times. Consequently, we identified a trade-off at $R = 10$, which is set as the default value in our experiments. More ablation studies on the hyperparameters can be found in Appendix B.
\par
\begin{figure}[tb]
	\centering
	\includegraphics[width=0.8\columnwidth]{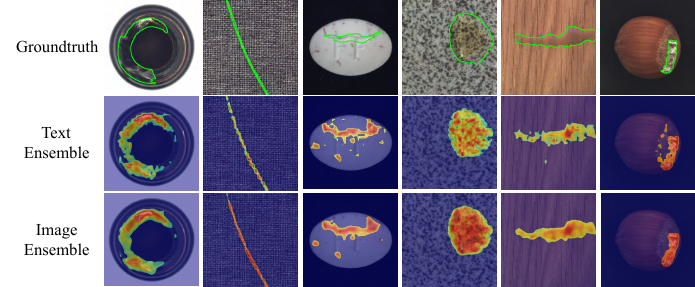}
	\caption{Visualization of different ensemble modes.}
	\label{Fig6}
\end{figure}

 
\textbf{Influence of the ensemble mode.} Given the text embeddings $\mathbf{t}_{b,r}^n$ and $\mathbf{t}_{b,r}^a$, we consider two ensemble modes: 1) Text ensemble, where text embeddings are averaged first and then aligned with image features; and 2) Image ensemble, where image features are aligned separately first, followed by averaging the anomaly scores or maps. Table \ref{Tab4} compares the ZSAD performance of these modes along with the average inference time per image. The results show that image ensemble outperforms text ensemble while maintaining comparable inference efficiency. As illustrated in Figure \ref{Fig6}, image ensemble produces more complete detection results with higher confidence in anomalous regions. Therefore, we adopt image ensemble as the default approach for generating the final ZSAD results.

\begin{table}[tb]
	\caption{Ablation on different ensemble mode}
\centering
\label{Tab4}
\renewcommand{\arraystretch}{1}
\resizebox{0.8\columnwidth}{!}
{
	\begin{tabular}{>{\centering\arraybackslash}p{2.7cm}*{4}{>{\centering\arraybackslash}p{1cm}}>{\centering\arraybackslash}p{1.2cm}}
		\toprule
		\multirow{2}{*}{Ensemble   Mode} & \multicolumn{2}{c}{Image-level} & \multicolumn{2}{c}{Pixel-level} & \multirow{2}{*}{\makecell[c]{Time \\ (ms)}} \\ \cmidrule(lr){2-3}  \cmidrule(lr){4-5}
		& AUROC           & AP            & AUROC           & AP            &                       \\  \midrule
		Text Ensemble                    & 92.0            & 96.3          & 91.1            & 47.6          & \textbf{436.6}                 \\
		Image Ensemble                   & \textbf{92.3}            & \textbf{96.7}          & \textbf{91.8}            & \textbf{48.3}          & 443.4       \\
		\bottomrule          
	\end{tabular}
}
\end{table}

%% file: sec/5_discussion.tex
\section{Conclusion}
This paper presents a novel ZSAD method, Bayes-PFL, which models text prompt spaces as a probabilistic distribution. A prompt flow module generates ISD and IAD to model context and state prompt spaces, capturing prompt uncertainty and enhancing generalization to new categories by regularizing the prompt space. Rich text prompts are generated by combining distribution sampling with prompt banks, improving ZSAD performance through an ensemble approach. Additionally, an RCA module aligns dynamic text embeddings with fine-grained image features. These designs enable Bayes-PFL to achieve state-of-the-art ZSAD performance on 15 publicly available industrial and medical datasets.

\section*{\textbf{Acknowledgement}}
This work is supported in part by the National Science and Technology Major Project of China under Grant 2022ZD0119402; in part by the National Natural Science Foundation of China under Grant No 62373350 and 62371179; in part by the Youth Innovation Promotion Association CAS (2023145); in part by the Beijing Nova Program 20240484687.

%% file: sec/X_suppl.tex
\clearpage
\setcounter{page}{1}
\appendix
\begin{center}
	\large \textbf{Appendix}
\end{center}
\par
This appendix includes the following five parts: 1) Implementation details of our Bayes-PFL and the introduction of the state-of-the-art approaches in Section \ref{secA}; 2) Additional experimental results, including ablation studies, and further analysis in Section \ref{secB}; 3) Introduction to 15 industrial and medical datasets in Section \ref{secC}; 4) Presentation of more detailed quantitative and qualitative results in Section \ref{secD}; 5) Limitations of our method in Section \ref{secE}. 

\section{Implementation Details and State-of-the-art Methods}  \label{secA}
\subsection{Implementation Details}
\textbf{Details of the model architecture.} 
In Bayes-PFL, ViT-L-14-336 pretrained in CLIP \cite{CLIP} is adopted as the default backbone, with its image encoder consisting of 24 transformer layers. Following previous work \cite{VAND, AnomalyCLIP, AdaCLIP}, patch-level features at equal intervals from the image encoder (i.e., the 6th, 12th, 18th, and 24th layers) are extracted for fine-grained anomaly segmentation. Additionally, all input images are resized to a fixed resolution of $518 \times 518$ before being fed into the image encoder. Prior to obtaining the patch-level image features $\mathbf{F}^I$, a single linear layer without bias is first applied to map the raw patch features into the joint text-image space. In the prompt flow module, the flow length $K$ is set to 10 by default. The networks $f_\mu$ and $f_\Sigma$ used to generate the initial distribution's mean $\mu(\boldsymbol{\xi})$ and covariance $\Sigma(\boldsymbol{\xi})$, have the same MLP architecture, as illustrated by the prompt flow module in Figure 2 of the main text. Each network consists of three linear layers, with hidden layer dimensions equal to the joint embedding dimension $C$. SoftPlus activation function \cite{Tanh} is applied between every two consecutive layers. In the prompt banks, the number of learnable prompts in each bank is set to 3. The length of learnable context vectors $P$ and learnable state vectors $Q$ are both set to 5. During the auxiliary training and inference stages, the number of Monte Carlo sampling interations $R$ is set to 1 and 10, respectively.
\par 
\begin{algorithm}[tb]
	\caption{Sampling with prompt flow module}
	\label{Alg1}
	\textbf{Input}: Condition input $\boldsymbol{\xi}$; Two neural networks with the same architecture $f_\mu$ and $f_\Sigma$; Invertible linear mapping $h_k, k=1,2,\cdots,K$; Number of Monte Carlo sampling interations $R$; \\
	\textbf{Output}: Approximate posterior distribution of the prompt space $q_K(\Phi_K)$; The sampled prompts $\boldsymbol{\varphi_r}, r= 1,2,\cdots,R$ from the distribution;
	\begin{algorithmic}[1] %[1] enables line numbers
		\STATE Generate the mean and covariance of the initial distribution $q_0(\Phi_0)$ conditioned on $\boldsymbol{\xi}$:
		$\mu(\boldsymbol{\xi}) = f_\mu(\boldsymbol{\xi})$, $\Sigma(\boldsymbol{\xi}) = f_\Sigma(\boldsymbol{\xi})$, and let $q_0{(\Phi_0)} = \mathcal{N}(\boldsymbol{\mu}(\boldsymbol{\xi}), \boldsymbol{\Sigma}(\boldsymbol{\xi}))$ \\
		\STATE Perform $K$ invertible linear mappings on the initial random vector $\Phi_0$ to obtain $\Phi_K =  h_K(h_{K-1}(\cdots h_1(\Phi_0)))$. Then the final prompt distribution $q_K(\Phi_K)$ is acquired from Equation (6) of the main text.
		\FOR{ Interation $r = 1: R$}
		\STATE Randomly sample $\epsilon_r$ from a standard normal distribution $\mathcal{N}(0, \mathbf{I})$; \\
		\STATE Generate the initial sampled prompt using the reparameterization method: $p_r = \mu(\boldsymbol{\xi}) + \epsilon_r \odot \Sigma(\boldsymbol{\xi})$  \\
		\STATE Generate the final sampled prompts from $q_K(\Phi_K)$: $\boldsymbol{\varphi_r} =  h_K(h_{K-1}(\cdots h_1(p_r)))$
		\ENDFOR
		\STATE \textbf{return} $q_K(\Phi_K), \boldsymbol{\varphi_r}$
	\end{algorithmic}
\end{algorithm}
\textbf{Details of sampling with prompt flow module.} The prompt flow module is used to approximate the Bayesian posterior distribution of the text prompt space using variational inference. As shown in Algorithm \ref{Alg1}, it learns to map a simple initial distribution $q_0(\Phi_0)$ through a series of invertible linear transformations to a more complex distribution $q_K(\Phi_K)$ that closely approximates the true posterior. However, sampling from the distribution using Monte Carlo methods is a discrete process, which prevents the prompt flow module from optimizing its parameters through backpropagation. Therefore, reparameterization technique \cite{VAE} is used, where two neural networks, $f_\mu$ and $f_\Sigma$, generate the mean $\mu(\boldsymbol{\xi})$ and covariance $\Sigma(\boldsymbol{\xi})$ of the initial normal distribution conditioned on $\boldsymbol{\xi}$, respectively. The symbol $\boldsymbol{\xi}$  represents either global image features or learnable free vectors. A single sample from the initial distribution is then acquired by computing: $p_r = \mu(\boldsymbol{\xi}) + \epsilon_r \odot \Sigma(\boldsymbol{\xi})$, where $\epsilon_r$ is obtained by randomly sampling from a standard normal distribution. The final sample $\boldsymbol{\varphi_r}$ can be obtained through $K$ invertible linear transformations as follows: $\boldsymbol{\varphi_r} =  h_K(h_{K-1}(\cdots h_1(p_r))), r = 1,2,\cdots, R$. 
\par
\textbf{Details of training and inference.} We conduct experiments using seven datasets from the industrial domain and eight datasets from the medical domain. The industrial dataset VisA \cite{VisA} is used as an auxiliary training set to fine-tune our Bayes-PFL. The resulting weights are directly applied to test on other industrial and medical datasets in a zero-shot manner. For VisA, the industrial dataset MVTec-AD \cite{MVTec} is adopted as the auxiliary training set. For the efficiency in auxiliary training phase, a single Monte Carlo sampling (i.e. $R = 1$) is used and the corresponding text embeddings are denoted as $\mathbf{Z}_{b,1}^t, b = 1,2,\dots, B$. At each training step, we randomly select one from the $B$ text embeddings to align with the image features. The final ZSAD result for each dataset is obtained by averaging the metrics of all categories contained within it. Note that during training, both the vanilla image and text encoders of CLIP are frozen, and only the parameters in newly added modules (e.g. the RCA module) are updated. This allows all of our experiments to be conducted on a single NVIDIA GeForce RTX 3090 with 24GB GPU memory. The optimizer used in our experiments is Adam \cite{Adam}, with an initial learning rate of 0.0001 and a batch size of 32. The model is trained for 20 epochs. We employ a cosine learning rate scheduler with warm-up, where the number of warm-up steps is set to 10\% of the total training steps. To reduce unnecessary computation, we also apply an early stopping strategy. Training is halted if the validation performance does not improve for a fixed number of consecutive epochs (patience=3). Five rounds of experiments are conducted using different random seeds, and the results are averaged to reduce experimental bias.
\par 
During the inference stage, $R = 10$ Monte Carlo samples are taken, and the corresponding text embeddings are denoted as $\mathbf{Z}_{b,r}^t, b=1,2,\cdots,B, r=1,2,\dots,R$. These $B\times R$ text embeddings are each aligned with the image features, and the resulting anomaly maps and anomaly scores are simply integrated using averaging operation to obtain the final results. Note that when aligning with the patch-level features, we additionally employ the proposed RCA module to refine the text embeddings to $\mathbf{F}_{b,r}^t$
\par 
\textbf{Details of the loss function.} In Equation (14) of the main text, we state that the auxiliary training loss is derived as the sum of the orthogonal loss $\mathcal{L}_{ort}$ and the prompt flow loss $\mathcal{L}_p$. For $\mathcal{L}_{ort}$, it enforces the diversity of different prompts in the prompt banks through orthogonal constraints and can be easily computed from Equation (13) in the main text. Here, we provide a further explanation of $\mathcal{L}_p$ as introduced in Equation (7) of the main text. For convenience of explanation, we copy it from the main text below: 
\begin{align}
	\mathcal{L}_p 
	&= E_{q_0(\Phi_0)}\left[\log q_0(\Phi_0) - \sum_{k=1}^{K}\log|1 + \mathbf{u}_k^T\phi(\Phi_k)|\right] \notag \\
	&\hspace{-2em}  - E_{q_0(\Phi_0)}[\log p(\Phi_K)] - E_{q_0(\Phi_0)}[\log p(D | \Phi_K)]
\end{align}
where $p = \mathcal{N}(0,\mathbf{I})$ and $q_0 = \mathcal{N}(\boldsymbol{\mu}(\boldsymbol{\xi}), \boldsymbol{\Sigma}(\boldsymbol{\xi}))$. The first two items can be easily obtained by calculating the log-likelihood for different sampled prompts under the assumption of a normal distribution. They ensure that the complex prompt distribution $q_K(\Phi_K)$ can be computed using a simple normal distribution $q_0(\Phi_0)$. For the third item, the goal is to maximize the log-likelihood of the auxiliary training data. This is approximated as the sum of the classification loss $\mathcal{L}_{cls}$ and the segmentation loss $\mathcal{L}_{seg}$:
\begin{align}
	- E_{q_0(\Phi_0)}[\log p(D | \Phi_K)] 
	&= \mathcal{L}_{cls} + \mathcal{L}_{seg} \notag \\
	&\hspace{-3em}=  \mathcal{L}_{cls} + (\mathcal{L}_{focal} + \mathcal{L}_{dice})
\end{align}
where $\mathcal{L}_{cls}$ is a simple cross-entropy loss for classification. For $\mathcal{L}_{seg}$, the sum of the Focal Loss $\mathcal{L}_{focal}$ \cite{Focal} and Dice Loss $\mathcal{L}_{dice}$ \cite{Dice} for semantic segmentation is adopted. Specifically, the Focal Loss is to tackle the class imbalance between the background and anomalous regions, such as in the VisA \cite{VisA} dataset, where the anomalous regions are significantly smaller than the background. It is computed as:
\begin{equation}
	\mathcal{L}_{focal} = - \frac{1}{N} \sum_{i=1}^{N} (1 - p_i)^\gamma \log(p_i)
\end{equation}
where $N$ is the number of pixels, $p_i$ is the probability predicted for the correct class. The symbol $\gamma$ is the focal factor, which adjusts the loss for easy and hard samples, and is set to 2 in this paper. Dice Loss quantifies the overlap between the predicted region and the ground truth, enabling the model to focus more on the anomalous areas. It is calculated as follows: 
\begin{equation}
	\mathcal{L}_{dice} = 1 - \frac{2\sum_{i}y_i\hat{y}_i}{\sum_{i}y_i+\sum_{i}\hat{y}_i}
\end{equation}
where $y_i$ and $\hat{y}_i$ represent the true label and predicted probability of the $i\text{-}th$ pixel in the image, respectively.

\begin{figure*}[t]
	\centering
	\begin{subfigure}[b]{0.49\textwidth}
		\centering
		\includegraphics[width=\textwidth]{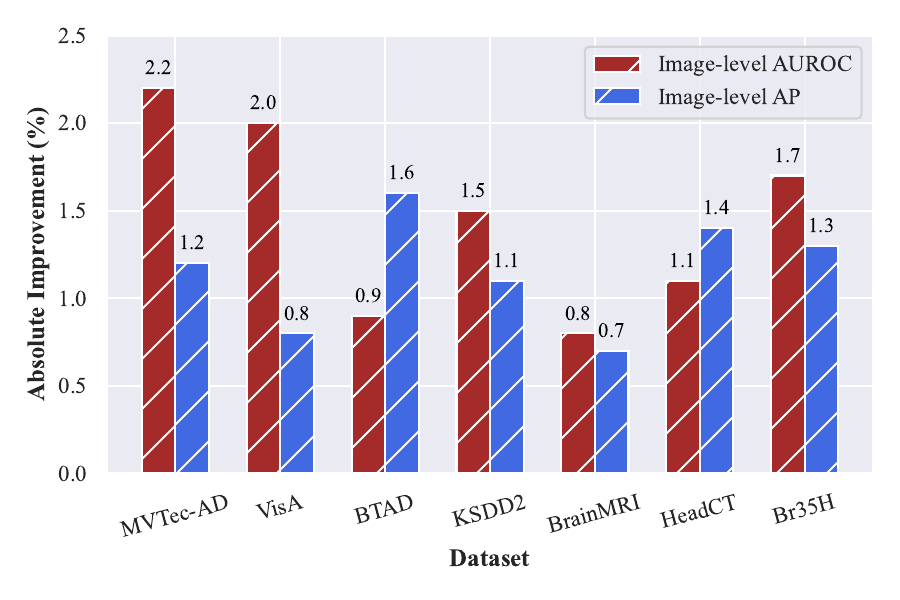}
		\caption{Absolute improvement on the image-level metrics}
		\label{fig2a}
	\end{subfigure}
	\hfill
	\begin{subfigure}[b]{0.49\textwidth}
		\centering
		\includegraphics[width=\textwidth]{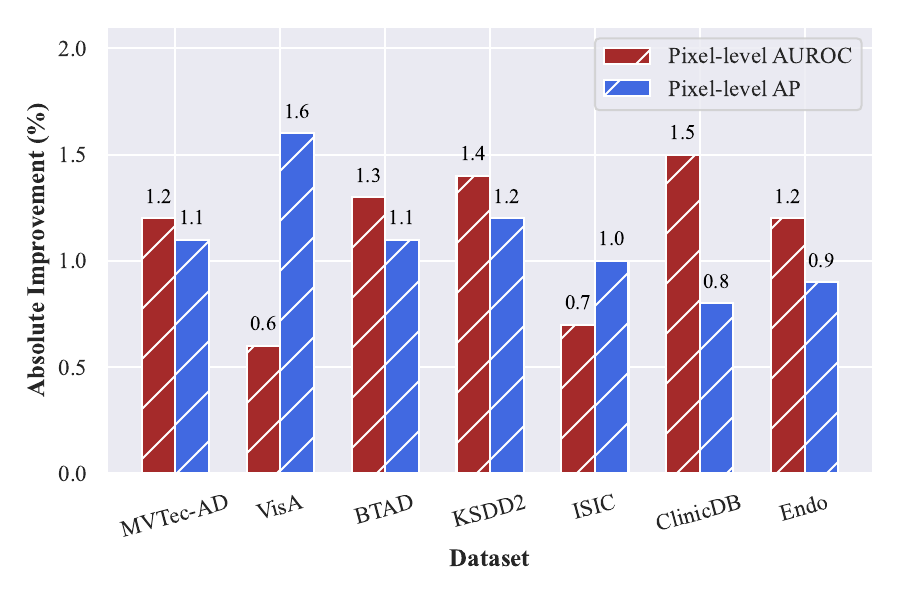}
		\caption{Absolute improvement on the pixel-level metrics}
		\label{fig2b}
	\end{subfigure}
	\caption{Absolute performance improvement of our prompt-flow-based distribution modeling over Gaussian modeling in ZSAD. (a) Improvement of image-level metrics. (b) Improvement of pixel-level metrics. All values being positive indicates that our prompt-flow-based distribution outperforms the Gaussian distribution in the text prompt space.}
	\label{fig2}
\end{figure*}
\subsection{State-of-the-art Methods}
\begin{itemize}
	
	\item \textbf{WinCLIP} \cite{WinCLIP} is one of the earliest works based on CLIP for the ZSAD task. Since the vanilla CLIP does not align text with fine-grained image features during pretraining, it addresses this limitation by dividing the input image into multiple sub-images using windows of varying scales. The final anomaly segmentation results are derived by harmoniously aggregating the classification outcomes of sub-images corresponding to the same spatial locations. In addition, a two-class text prompt design method, named Compositional Prompt Ensemble, is proposed and has been widely adopted in subsequent works.
	
	\item \textbf{APRIL-GAN} \cite{VAND} adopts the handcrafted textual prompt design strategy from WinCLIP. However, for aligning textual and visual features, it introduces a linear adapter layer to project fine-grained patch features into a joint embedding space. After training on an auxiliary dataset, it can directly generalize to novel categories. 
	
	\item \textbf{CLIP-AD} \cite{CLIPAD} builds upon the modality alignment design of APRIL-GAN, continuing to use a linear adapter to project patch-level image features. The key difference is that it incorporates a feature surgery strategy to further address the issues of opposite predictions and irrelevant highlights. 
	
	\item \textbf{AnomalyCLIP} \cite{AnomalyCLIP} employs a prompt-optimization-based text design strategy. By training on an auxiliary dataset, it learns object-agnostic text prompts that can be directly transferred to unseen object categories. 
	
	\item \textbf{AdaCLIP} \cite{AdaCLIP} introduces a hybrid prompt mechanism that integrates both dynamic and static prompts, embedding them into the text and image encoder layers. By incorporating visual prompts, the output text embeddings are able to dynamically adapt to the input image, thereby enhancing generalization performance.
\end{itemize}
\par
Since the official code for WinCLIP has not been released, we use the reproduced code from \cite{AnomalyCLIP}. For APRIL-GAN, CLIP-AD, AnomalyCLIP, and AdaCLIP, we re-trained the models using the official code, maintaining the same backbone, input image resolution, and experimental settings (training on the VisA dataset and testing on other datasets) as those used in our Bayes-PFL. This ensures the fairness of the comparison between our Bayes-PFL and other state-of-the-art (SOTA) methods.
\begin{table}[tp]
	\caption{The ZSAD performance when the single prompt (SP) and prompt ensemble (PE) are applied separately to the training and inference stages of the APRIL-GAN \cite{VAND}. The best results are shown in \textbf{bold}.}
	\centering
	\label{B1}
	\renewcommand{\arraystretch}{1.2}
	\resizebox{1\columnwidth}{!}
	{
		\begin{tabular}{*{4}{>{\centering\arraybackslash}p{0.5cm}}*{4}{>{\centering\arraybackslash}p{1.1cm}}}
			\toprule
			\multicolumn{2}{c}{Training} & \multicolumn{2}{c}{Inference} & \multicolumn{2}{c}{Image-level} & \multicolumn{2}{c}{Pixel-level} \\  \cmidrule(lr){1-2} \cmidrule(lr){3-4}  \cmidrule(lr){5-6} \cmidrule(lr){7-8}
			SP            & PE           & SP            & PE            & AUROC           & AP            & AUROC           & AP            \\ \midrule
			\ding{52}             &      \textcolor{gray}{\ding{56}}        & \ding{52}             &       \textcolor{gray}{\ding{56}}        & 81.7            & 91.6          & 86.8            & 37.4          \\
			\ding{52}             &       \textcolor{gray}{\ding{56}}       &      \textcolor{gray}{\ding{56}}         & \ding{52}             & 85.9            & 93.0          & 87.1            & 37.6          \\
			\textcolor{gray}{\ding{56}}	& \ding{52}            & \ding{52}             &       \textcolor{gray}{\ding{56}}        & 81.8            & 91.6          & 86.9            & 40.5          \\
			\textcolor{gray}{\ding{56}}	& \ding{52}            &       \textcolor{gray}{\ding{56}}        & \ding{52}             & \textbf{86.1}            & \textbf{93.5}          & \textbf{87.6 }           & \textbf{40.8 }        \\  \bottomrule
		\end{tabular}
	}
\end{table}
\begin{figure*}[tp]
	\centering
	\begin{subfigure}[b]{0.49\textwidth}
		\centering
		\includegraphics[width=\textwidth]{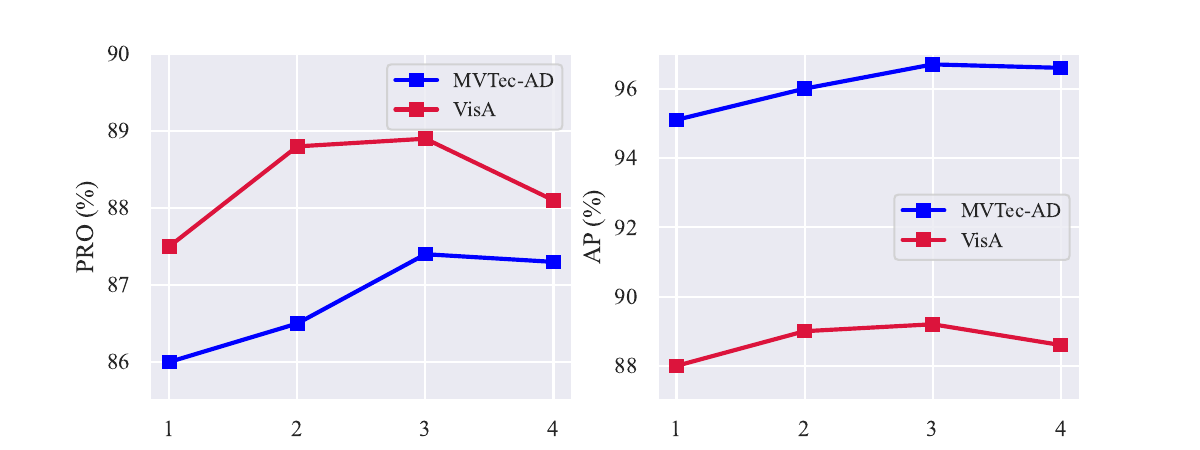}
		\caption{$B$}
		\label{fig1a}
	\end{subfigure}
	\hfill
	\begin{subfigure}[b]{0.49\textwidth}
		\centering
		\includegraphics[width=\textwidth]{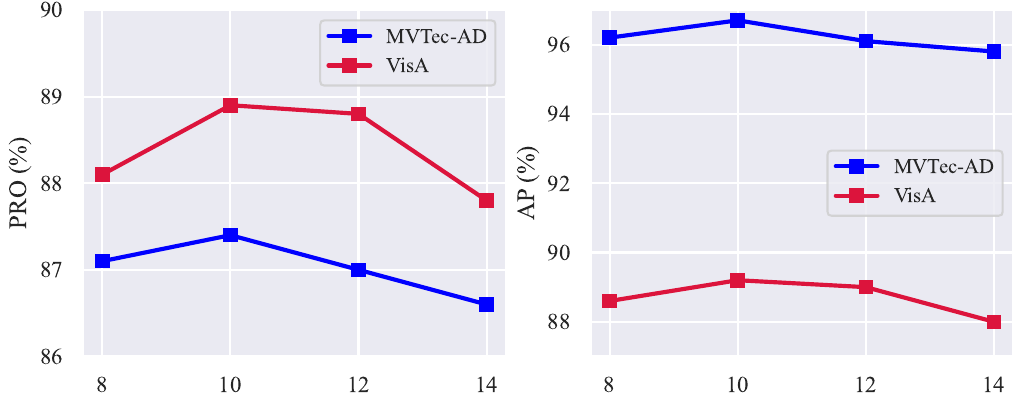}
		\caption{$K$}
		\label{fig1b}
	\end{subfigure}
	\vfill
	\begin{subfigure}[b]{0.49\textwidth}
		\centering
		\includegraphics[width=\textwidth]{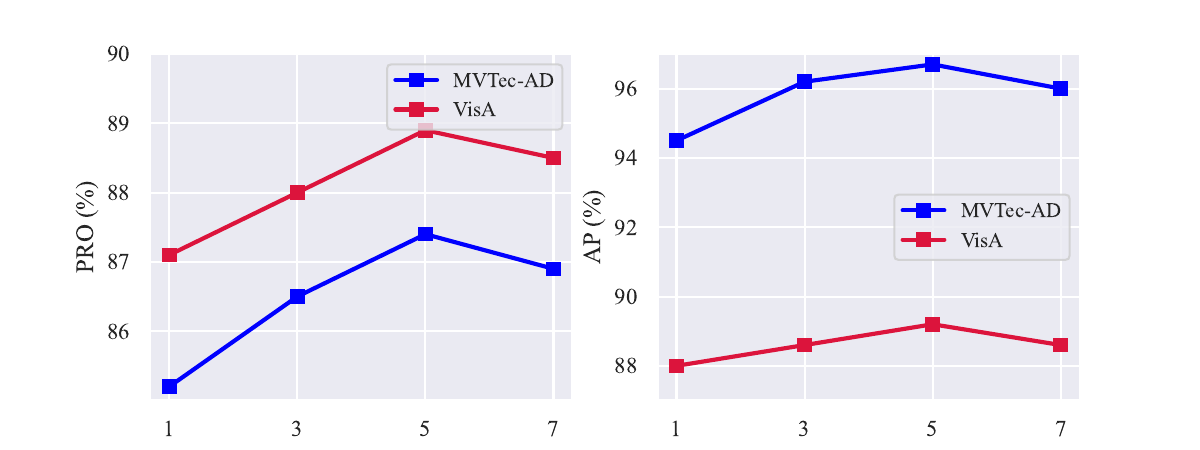}
		\caption{$P$}
		\label{fig1c}
	\end{subfigure}
	\hfill
	\begin{subfigure}[b]{0.49\textwidth}
		\centering
		\includegraphics[width=\textwidth]{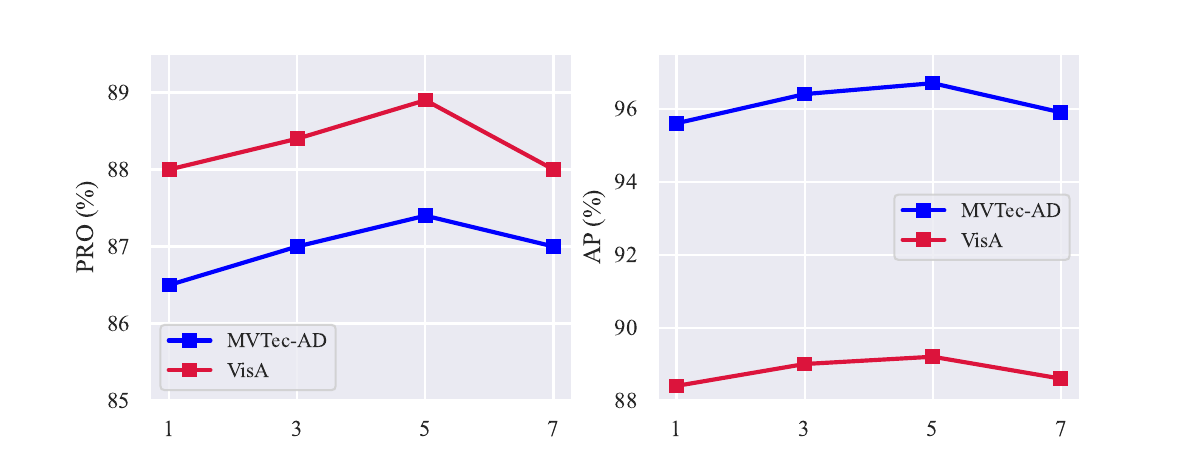}
		\caption{$Q$}
		\label{fig1d}
	\end{subfigure}
	\caption{(a) Ablation on the number of learnable prompts in each prompt bank $B$. (b) Ablation on the length of flow in the prompt flow module $K$. (c) Ablation on the length of learnable context vectors $P$. (d) Ablation on the length of learnable state vectors $Q$. Pixel/image-level (PRO, AP) performances are shown on the left and right sides of each subplot, respectively.}
	\label{fig1}
\end{figure*}
\section{Additional Results and Analysis}   \label{secB}
\subsection{Motivation}
In Bayes-PFL, the text prompt space is modeled as a learnable distribution, and Monte Carlo sampling is used to sample from it to cover the prompt space. \textbf{A key question arises: what motivates this approach?} Our idea originally stemmed from the observation of using different types of manually designed prompts in the training and inference stages of APRIL-GAN \cite{VAND}: single prompt (SP) or prompt ensemble (PE). For SP, the input text prompts are fixed, designed as \textit{a photo of a perfect [class]} and \textit{a photo of a damaged [class]} for normal and abnormal cases, respectively. For PE,  the prompt templates and state words from the original paper are utilized to generate a large number of text prompts through combinations, as detailed in Section 2.2 of the main text. The experimental results, presented in Table \ref{B1}, reveal that using SP during both training and inference yields the worst ZSAD performance. Additionally, employing PE consistently outperforms the cases where it is not used, demonstrating its effectiveness.
\par
\textbf{This inspires us that:} 1) using a single, fixed text prompt during training and inference hinders the model's generalization to novel categories; and 2) a richer and more diverse text prompt space can improve the model's ZSAD performance through the ensemble strategy. However, the number of manually designed text prompts is inherently limited, and their creation relies on expert knowledge, often requiring extensive trial and error. This observation motivates us to enhance the model's generalization performance on novel categories by learning a distribution over the prompt space, rather than relying solely on manually crafted prompts.

\subsection{Comparison of Different Distribution Types in the Text Prompt Sapce}
In Bayes-PFL, the prompt flow module is designed to approximate the true posterior distribution of the prompt space. Specifically, Image-specific distribution (ISD) and Image-agnostic distribution (IAD) are used to model the context and state text prompt spaces, respectively. This design shifts the focus away from the specific form of the distribution, allowing us to directly learn how to transform a simple normal distribution into the target distribution in a data-driven manner using the auxiliary dataset.
\par 
\textbf{Is the distribution based on prompt flow better than other fixed-form distributions (e.g. Gaussian distributions)?}  To investigate this, we model the text prompt space as a Gaussian distribution and compare it with our prompt-flow-based method. Specifically, the invertible linear transformation layer $h_k$ in the prompt flow module is removed, and the initial distribution  $q_0 = \mathcal{N}(\boldsymbol{\mu}(\boldsymbol{\xi}), \boldsymbol{\Sigma}(\boldsymbol{\xi}))$ is directly used as the variational distribution to approximate the true posterior distribution of the prompt space.
By substituting $q_0$ for $q_\gamma$ in Equation (5) of the main text, the new loss, which replaces $\mathcal{L}_p$, is computed as:
\begin{equation}
	\mathcal{L}_p' = D_{KL} [q_0(\Phi) || p(\Phi)] - E_{q_0(\Phi)}[\log p(D|\Phi)]
\end{equation}
where $D_{KL}$ denotes Kullback-Leibler divergence. The second term of Equation (A.5) is still approximated by the sum of classification and segmentation losses, which is similar to Bayes-PFL. 
\par 
Figure \ref{fig2} illustrates the absolute performance improvement of our prompt-flow-based distribution modeling over Gaussian modeling in the text prompt space. Seven different datasets from industrial or medical domains are used to evaluate the two distinct distribution types. It can be observed that the image-level and pixel-level metrics across all datasets are positive, indicating that our method outperforms Gaussian modeling. This is attributed to our Bayes-PFL's ability to adaptively learn arbitrary complex distributions for the prompt space, rather than being constrained to a fixed Gaussian distribution.
\subsection{Additional Ablations}
In this subsection, we perform additional ablation studies on hyperparameters as well as backbone and resolution settings.
\par 
\textbf{Ablation on hyperparameters.} As shown in Figure \ref{fig1}, ablation experiments on hyperparameters are conducted on the MVTec-AD and VisA datasets, including (a) the number of learnable prompts in each prompt bank $B$, (b) the length of the flow in the prompt flow module $K$, (c) the length of learnable context vectors $P$, and (d) the length of learnable state vectors $Q$. 
\par 
As observed in Figure \ref{fig1}(a), the model achieves optimal ZSAD performance when the number of learnable prompts in the prompt bank is set to 3. This is because our auxiliary training data is relatively simple. Under the orthogonal loss constraint, three distinct learnable prompts, when fused with the distribution sampling results, are sufficient to capture anomalous semantics in the training set and generalize to novel categories. Figure \ref{fig1}(b) demonstrates that the model attains the highest pixel-level PRO and image-level AP when the length of the prompt flow, $K$, is set to 10 across both datasets. A flow that is too short fails to map the initial distribution to the approximate posterior of the prompt space. Conversely, when $K$ becomes too large, the generalization performance declines. This is due to the increased number of invertible linear transformations, leading to an overly complex learned prompt distribution that hinders transferability to new data domains. For the learnable context vector length $P$ in Figure \ref{fig1}(c) and the state vector length $Q$ in Figure \ref{fig1}(d), the model's performance first improves and then declines, with optimal performance observed when the length is set to 5. These learnable vectors function as biases when fused with the prompts sampled from the distribution, making the final generated prompts closer to the true word embedding space. A small vector length is insufficient to capture the semantics of context and state, while a large length introduces redundant information, which also hinders generalization.
\par
\begin{figure*}[t]
	\centering
	\begin{subfigure}[b]{0.49\textwidth}
		\centering
		\includegraphics[width=\textwidth]{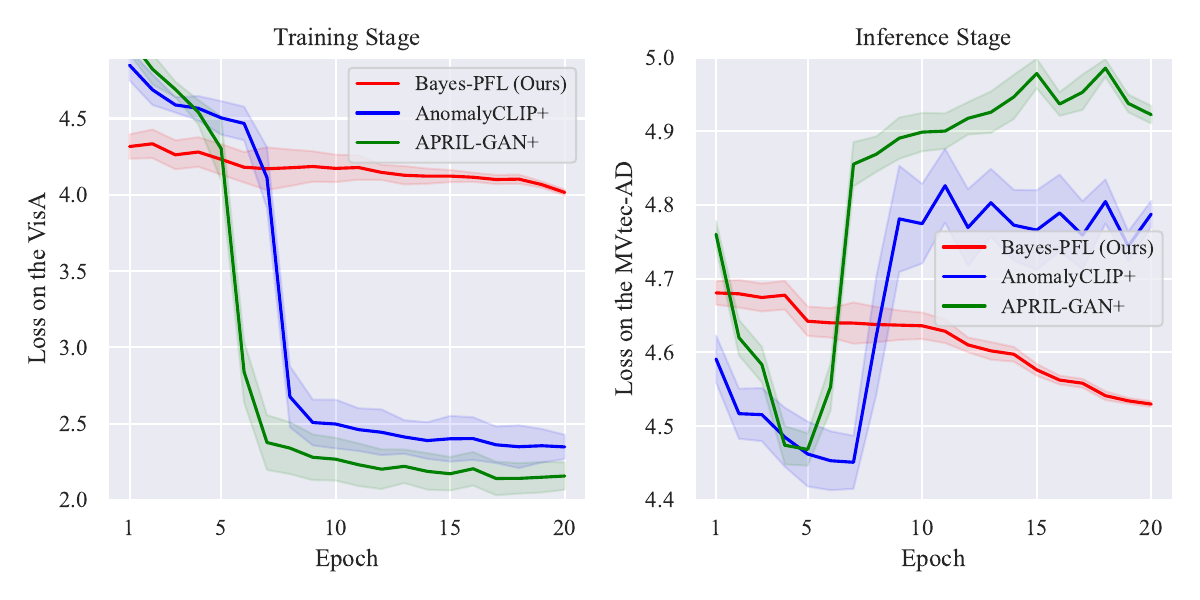}
		\caption{\textbf{Left}: Loss on the VisA during the training stage. \textbf{Right}: Loss on the MVTec-AD during the inference stage}
		\label{fig3a}
	\end{subfigure}
	\hfill
	\begin{subfigure}[b]{0.49\textwidth}
		\centering
		\includegraphics[width=\textwidth]{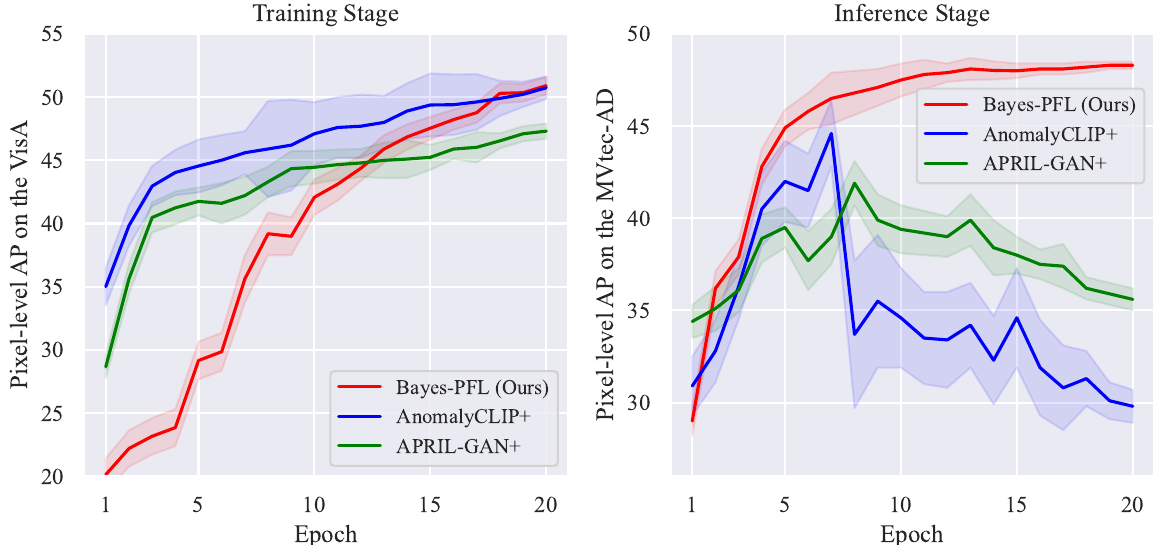}
		\caption{\textbf{Left}: Pixel-level AP on the VisA during the training stage. \textbf{Right}: Pixel-level AP on the MVTec-AD during the inference stage}
		\label{fig3b}
	\end{subfigure}
	\caption{(a) The variations in loss across different methods during training and inference stages. (b) The variations in pixel-level AP across different methods during training and inference stages. }
	\label{fig3_epoch}
\end{figure*}
\textbf{Influence of different backbones and resolutions.} In Table \ref{B3}, we analyze the effect of various pre-trained CLIP backbones and input image resolutions on the model's performance. The results indicate that larger backbones and appropriately higher input resolutions lead to more precise pixel-level segmentation. Notably, the ViT-L-14-336 backbone achieves the best ZSAD performance when the input image resolution is set to $518^2$. Consequently, we adopt this configuration as the default setting for our experiments. To ensure a fair comparison, we maintain consistent settings for other methods, including \cite{VAND, CLIPAD, AnomalyCLIP, AdaCLIP}.

\begin{table}[t]
	\caption{Ablation on different bankbone and input image size.}
	\centering
	\label{B3}
	\renewcommand{\arraystretch}{1}
	\resizebox{0.9\columnwidth}{!}
	{
		\begin{tabular}{>{\centering\arraybackslash}p{2cm}>{\centering\arraybackslash}p{0.7cm}*{4}{>{\centering\arraybackslash}p{1cm}}}
			\toprule
			\multirow{2}{*}{Backbone} & \multirow{2}{*}{Size} & \multicolumn{2}{c}{Image-level} & \multicolumn{2}{c}{Pixel-level} \\   \cmidrule(lr){3-4}  \cmidrule(lr){5-6}
			&                             & AUROC           & AP            & AUROC           & AP            \\  \midrule
			ViT-B-16-224              & $336^2$                         & 87.0            & 93.6          & 90.7            & 39.9          \\
			ViT-L-14-224              & $336^2$                         & 92.1            & 96.0          & 91.6            & 44.1          \\
			ViT-L-14-224              & $518^2$                         & 89.7            & 95.3          & 91.5            & 44.6          \\
			ViT-L-14-336              & $336^2$                         & 91.7            & 96.2          & 90.9            & 44.1          \\   \rowcolor{lightgreen}
			ViT-L-14-336              & $518^2$                         & 92.3            & 96.7          & 91.8            & 48.3          \\
			ViT-L-14-336              & $700^2$                         & 90.1            & 94.3          & 90.2            & 46.4           \\ \bottomrule    
		\end{tabular}
	}
\end{table}

\subsection{Additional Analysis}
In this subsection, we evaluate the generalization performance and inference efficiency of various models. Furthermore, the text embedding is visualized to provide deeper insights into the proposed Bayes-PFL. 
\par 
\textbf{Analysis of generalization performance.} In Section 4.2 of the main text, we compared the ZSAD performance of two improved methods, APRIL-GAN+ and AnomalyCLIP+, as well as our Bayes-PFL, across different training epochs. Experimental results show that the pixel-level AP of APRIL-GAN+ and AnomalyCLIP+ on the test set declines after reaching its peak. Here, we further investigate this phenomenon.
\par 
Figure \ref{fig3_epoch} shows the changes in both loss and pixel-level AP during the training and inference stages across different training epochs. APRIL-GAN+ and AnomalyCLIP+ demonstrate a consistent decrease in training loss, while the test loss decreases initially and then increases. In the training set, the AP steadily improves, whereas in the test set, the AP rises to a peak before exhibiting a downward trend. These trends suggest that both models gradually overfit the auxiliary training data, resulting in reduced generalization performance on unseen categories. In contrast, Bayes-PFL shows a gradually decreasing trend in test loss, with AP slowly increasing and eventually stabilizing during the inference stage. This indicates that our text prompt distributional strategy effectively mitigates overfitting to the training data, resulting in stronger zero-shot transfer capabilities.
\par 
\textbf{Analysis of inference efficiency.} Table \ref{Table3} compares the ZSAD performance of different methods, along with the maximum GPU consumption per image during inference and the average inference time per image on the MVTec-AD dataset. The proposed Bayes-PFL achieves faster inference speed than WinCLIP \cite{WinCLIP}, but is slower than other methods. This is primarily due to the additional time required for sampling multiple text prompts from the distribution and performing inference. However, the ZSAD performance of Bayes-PFL significantly outperforms other methods, and this trade-off between inference speed and performance is deemed acceptable given its stronger generalization capability. Moreover, compared to APRIL-GAN and AdaCLIP, Bayes-PFL does not incur additional GPU memory consumption. This is because it avoids storing fixed text embeddings for all categories in memory, instead dynamically updating them based on the input image. The dynamic adjustment of text based on the input image enhances the model's anomaly detection capability for unseen categories to some extent.
\begin{table}[tp]
	\caption{Comparison of average inference time and maximum GPU cost on MVTec-AD. The best results are shown in \textbf{bold}.}
	\centering
	\label{Table3}
	\renewcommand{\arraystretch}{1}
	\resizebox{1\columnwidth}{!}
	{
		\begin{tabular}{>{\centering\arraybackslash}p{2cm}*{4}{>{\centering\arraybackslash}p{1cm}}>{\centering\arraybackslash}p{1.1cm}>{\centering\arraybackslash}p{1.1cm}}
			\toprule
			\multirow{2}{*}{Method} & \multicolumn{2}{c}{Image-level} & \multicolumn{2}{c}{Pixel-level} & \multirow{2}{*}{\makecell[c]{GPU \\ cost (GB)}} & \multirow{2}{*}{\makecell[c]{Time \\ (ms)}} \\   \cmidrule(lr){2-3}  \cmidrule(lr){4-5} 
			& AUROC           & AP            & AUROC           & AP            &                            &                       \\   \midrule
			WinCLIP                 & 91.8            & 95.1          & 85.1            & 18.0          & \textbf{2.0 }                       & 840.0                 \\
			APRIL-GAN               & 86.1            & 93.5          & 87.6            & 40.8          & 3.3                        & \textbf{105.0}                 \\
			CLIP-AD                 & 89.8            & 95.3          & 89.8            & 40.0          & 3.4                        & 115.2                 \\
			AnomalyCLIP             & 91.5            & 96.2          & 91.1            & 34.5          & 2.7                        & 131.2                 \\
			AdaCLIP                 & 92.0            & 96.4          & 86.8            & 38.1          & 3.3                        & 183.4                 \\  \rowcolor{lightgreen}
			Bayes-PFL               & \textbf{92.3}            & \textbf{96.7}          & \textbf{91.8}            & \textbf{48.3}          & 3.3                        & 388.5                \\ \bottomrule        
		\end{tabular}
	}
\end{table}
\begin{figure}[t]
	\centering
	\includegraphics[width=1\columnwidth]{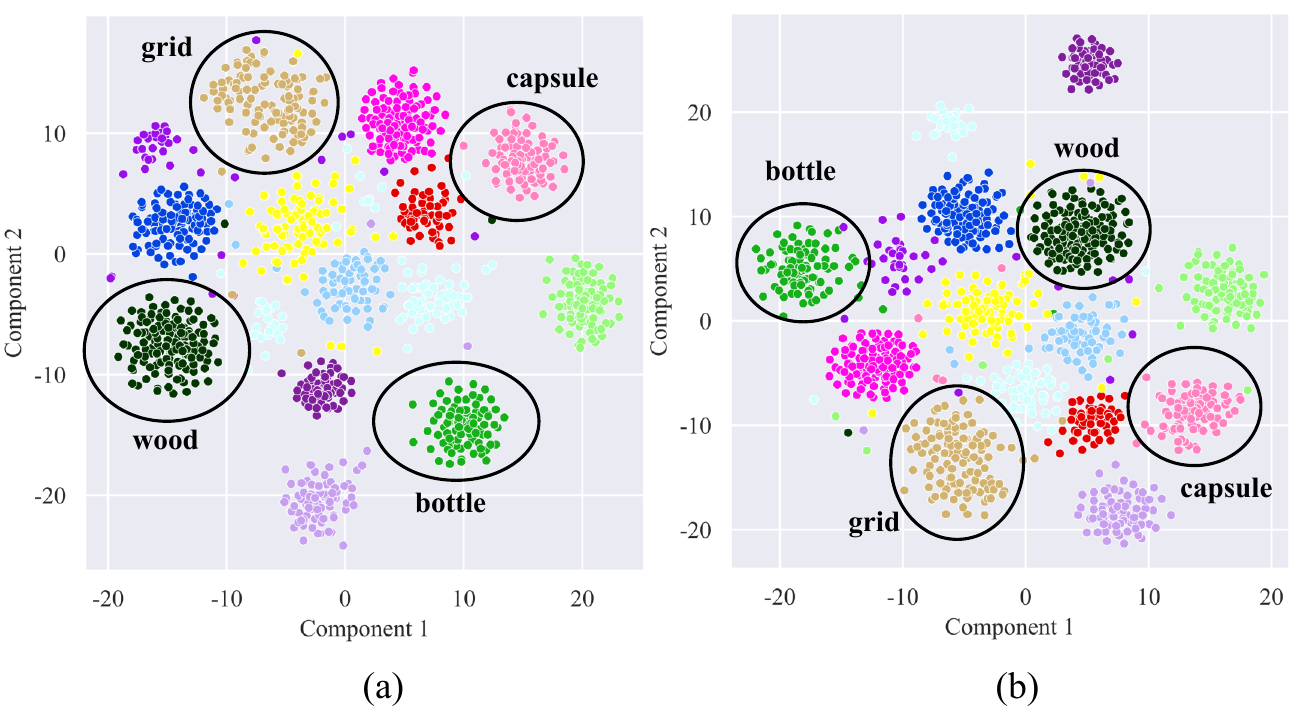}
	\caption{The t-SNE visualization of text embeddings with two Monte Carlo samplings on the MVTec-AD dataset. (a) The first sampling. (b) The second sampling.}
	\label{fig4}
\end{figure}
\par 
\textbf{Analysis of the text embedding space.} In Bayes-PFL, we utilize the Image-Specific Distribution (ISD) and the Image-Agnostic Distribution (IAD) to model context words and state words in the text prompt space, respectively. Therefore, direct visualization of the prompt space is challenging. To address this, we employ t-SNE for dimensionality reduction to visualize the output of the text encoder, i.e., the text embeddings $\mathbf{Z}_{b,r}^t$. 
Figure \ref{fig4} presents the visualization results of text embeddings corresponding to abnormal prompts under two Monte Carlo sampling instances on the MVTec-AD dataset. Note that due to the introduction of ISD, each test image corresponds to a single data point in the picture. Two phenomena can be observed: 1) Text embeddings of the same category are clustered together, but differences still exist among individual image; 2) Different Monte Carlo sampling instances lead to corresponding changes in the text embeddings. 
\par 
These observations align with our expectations. First, the introduction of visual features in ISD results in distinct prompt distributions for each test image, leading to differences in the text embeddings of individuals from the same category. However, due to the effective integration of visual semantics into the text prompts, text embeddings of the same category exhibit commonalities, causing them to cluster together. Second, the purpose of sampling from different distributions is to cover the prompt space as comprehensively as possible, thereby enhancing the generalization capability of the prompts for unseen categories. Consequently, different Monte Carlo sampling instances contribute to more diverse and enriched text embeddings, which improves the ZSAD performance.

\section{Datasets}   \label{secC}
In this section, we provide a brief introduction to the 15 datasets used in this work.
\subsection{Industrial Domain}

\begin{itemize}
	\item \textbf{MVTec-AD} \cite{MVTec} is specifically designed for industrial anomaly detection, consisting of 15 different categories (e.g., bottle, wood). In this work, we use only its labeled test set, which includes 467 normal images and 1,258 anomalous images. It also includes data from both texture and object types, making it more comprehensive, and is widely used in our ablation experiments. 
	
	\item \textbf{VisA} \cite{VisA} is a challenging industrial dataset that includes 12 categories (e.g., candle, capsules), all of which are object types. Its test set contains 962 normal images and 1,200 anomalous images, and is primarily used for auxiliary training in this study. Due to the small size of the anomalous regions relative to the background, it presents a significant challenge for both pixel-level classification and image-level segmentation. 
	
	\item \textbf{BTAD} \cite{BTAD} includes three categories, all of which are object types, with resolutions ranging from 600 to 1600. The dataset contains 451 normal images and 290 anomalous images, used to evaluate the ZSAD performance.
	
	\item \textbf{KSDD2} \cite{KSDD2} is a dataset designed for industrial defect detection. It includes 2,085 normal images and 246 abnormal images in the original training set, as well as 894 normal images and 110 abnormal images in the original test set. The image dimensions are similar, approximately 230 pixels in width and 630 pixels in height. In this work, we reconstruct the dataset for ZSAD. Specifically, all 356 abnormal images from the original training and test sets, along with an equal number of randomly selected normal images, are combined to form a new dataset. Note that this differs from the dataset processing approach in VCP-CLIP \cite{VCP}, where only all anomalous samples from the test set are used for zero-shot anomaly segmentation evaluation.
	
	\item \textbf{RSDD} \cite{RSDD} is a dataset for rail surface defect detection, consisting of two categories, both of which are object types. Its original dataset contains a total of 195 high-resolution abnormal images. To adapt it for the ZSAD task, the original images are cropped to generate 387 normal images and 387 abnormal images, each with a resolution of $512 \times 512$. Note that this differs from the dataset processing approach in VCP-CLIP \cite{VCP}, where only all anomalous samples from the test set are used for zero-shot anomaly segmentation evaluation.
	
	\item \textbf{DAGM} \cite{DAGM} is a texture dataset designed for weakly supervised anomaly detection, consisting of 10 categories. It contains 6,996 normal images and 1,054 abnormal images. Since the original pixel-level annotations are weak labels in the form of ellipses, we manually re-annotate the DAGM dataset for anomaly segmentation. 
	
	\item \textbf{DTD-Synthetic} \cite{DTD} is a synthetic dataset designed for texture anomaly detection, comprising 12 categories. It contains 357 normal images and 947 anomalous images. 
\end{itemize}

\subsection{Medical Domain}
\begin{itemize}
	
	\item \textbf{HeadCT} \cite{HeadCT} is a dataset for head CT scan analysis, containing images of brain scans with various conditions, such as hemorrhages and tumors, intended for anomaly detection. It includes a total of 100 normal images and 100 anomalous images, with image-level labels only. Therefore, it is used exclusively for the anomaly classification task. In this work, we directly adopt the dataset curated by AdaCLIP \cite{AdaCLIP}.
	
	\item \textbf{BrainMRI} \cite{BrainMRI} is a dataset for brain MRI analysis, comprising images of both healthy and abnormal brain scans, including conditions like tumors and lesions. It contains 98 anomalous images and 155 normal images, with only image-level labels. In this work, we directly adopt the dataset curated by AdaCLIP \cite{AdaCLIP}.
	
	\item \textbf{Br35H} \cite{Br35H} is a dataset for brain tumor detection in MRI images, containing 1500 normal images and 1500 anomalous images. Since it only provides image-level labels, it is used solely for anomaly classification task. In this work, we directly adopt the dataset curated by AdaCLIP \cite{AdaCLIP}.
	
	\item \textbf{ISIC} \cite{ISIC} is a dataset for skin lesion analysis in endoscopy images, containing a large collection of dermoscopic images, each labeled as either melanoma or non-melanoma lesions. It includes 379 anomalous images with pixel-level annotations, making it suitable only for anomaly segmentation task. In this work, we directly adopt the dataset curated by AdaCLIP \cite{AdaCLIP}.
	
	\item \textbf{CVC-ColonDB} \cite{ColonDB} is a dataset for colorectal cancer detection in endoscopy images, containing colonoscopy images with labeled polyps. It includes a total of 380 anomalous images for colon polyp detection, with pixel-level annotations. Consequently, it is used for anomaly segmentation tasks in the medical domain in this work. In this work, we directly adopt the dataset curated by AdaCLIP \cite{AdaCLIP}.
	
	\item \textbf{CVC-ClinicDB} \cite{ClinicDB} is similar to CVC-ColonDB, containing 612 anomalous images with pixel-level annotations. Therefore, it is also used exclusively for anomaly segmentation tasks. In this work, we directly adopt the dataset curated by AdaCLIP \cite{AdaCLIP}.
	
	\item \textbf{Endo} \cite{Endo} is another dataset similar to CVC-ColonDB, comprising 200 anomalous images with pixel-level annotations. While it is also used for colon polyp detection, differences in image acquisition devices and environments introduce a certain domain gap compared to other datasets. In this work, we directly adopt the dataset curated by AnomalyCLIP \cite{AnomalyCLIP}.
	
	\item \textbf{Kvasir} \cite{kvasir}  is a larger medical dataset used for colon polyp detection in endoscopy images. It contains 1,000 anomalous images with pixel-level annotations and is used for anomaly segmentation task in the medical domain in this work. In this work, we directly adopt the dataset curated by AnomalyCLIP \cite{AnomalyCLIP}.
\end{itemize}

\section{Detailed ZSAD results}  \label{secD}
\begin{figure*}[tb]
	\centering
	\includegraphics[width=1.8\columnwidth]{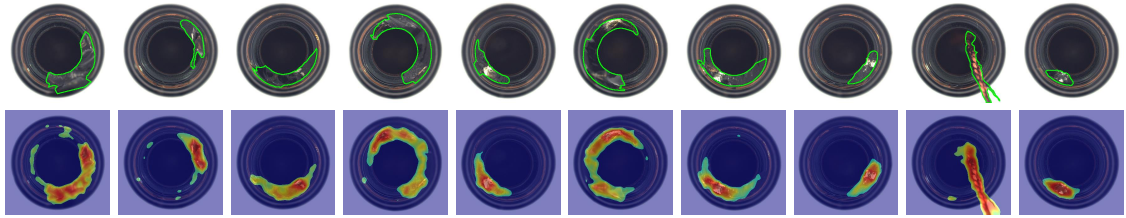}
	\caption{Visualization of segmentation results for the bottle category on MVTec-AD. The first row represents the input images, with green outlines indicating the ground truth regions. The second row shows the results of anomaly segmentation.}
	\label{fig_r5}
\end{figure*}

\begin{figure*}[tb]
	\centering
	\includegraphics[width=1.8\columnwidth]{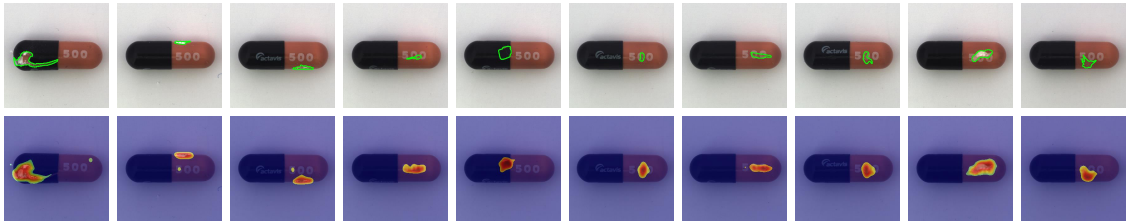}
	\caption{Visualization of segmentation results for the capsule category in MVTec-AD. The first row represents the input images, with green outlines indicating the ground truth regions. The second row shows the results of anomaly segmentation.}
	\label{fig_r6}
\end{figure*}

\begin{figure*}[tb]
	\centering
	\includegraphics[width=1.8\columnwidth]{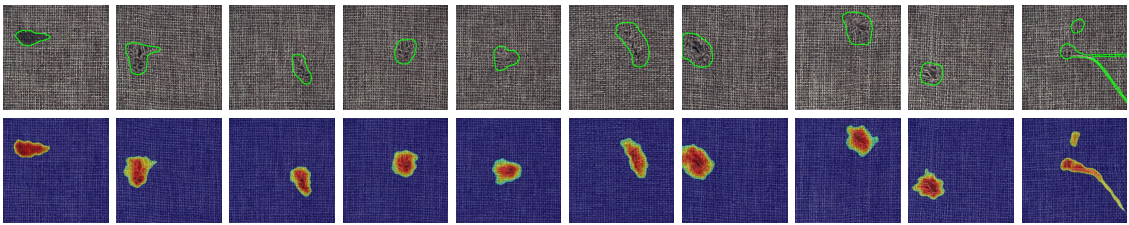}
	\caption{Visualization of segmentation results for the carpet category in MVTec-AD. The first row represents the input images, with green outlines indicating the ground truth regions. The second row shows the results of anomaly segmentation.}
	\label{fig_r7}
\end{figure*}

\begin{figure*}[tb]
	\centering
	\includegraphics[width=1.8\columnwidth]{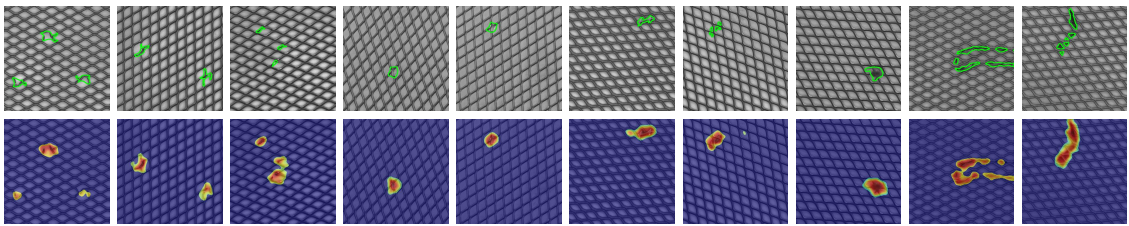}
	\caption{Visualization of segmentation results for the grid category in MVTec-AD. The first row represents the input images, with green outlines indicating the ground truth regions. The second row shows the results of anomaly segmentation.}
	\label{fig_r8}
\end{figure*}

\begin{figure*}[tb]
	\centering
	\includegraphics[width=1.8\columnwidth]{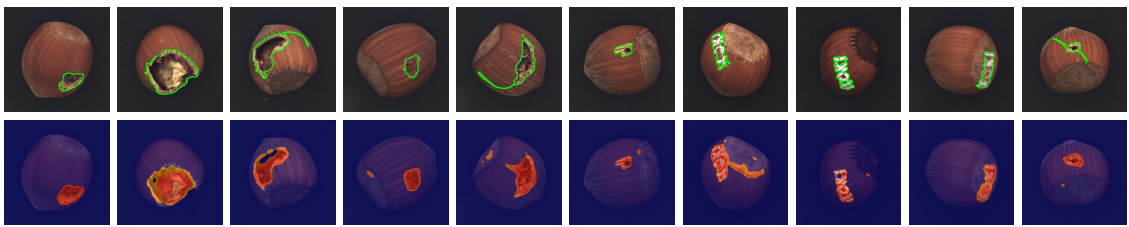}
	\caption{Visualization of segmentation results for the hazelnut category in MVTec-AD. The first row represents the input images, with green outlines indicating the ground truth regions. The second row shows the results of anomaly segmentation.}
	\label{fig_r9}
\end{figure*}

\begin{figure*}[tb]
	\centering
	\includegraphics[width=1.8\columnwidth]{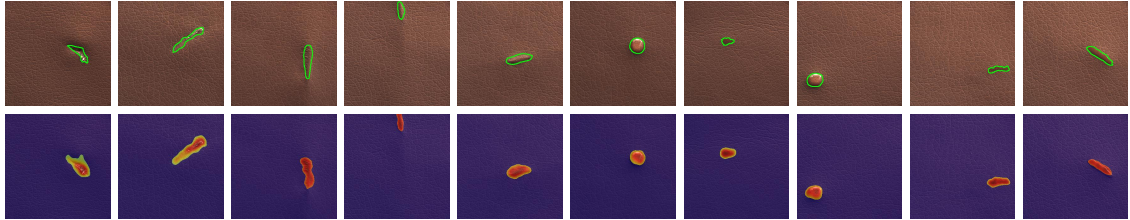}
	\caption{Visualization of segmentation results for the leather category in MVTec-AD. The first row represents the input images, with green outlines indicating the ground truth regions. The second row shows the results of anomaly segmentation.}
	\label{fig_r10}
\end{figure*}

\begin{figure*}[tb]
	\centering
	\includegraphics[width=1.8\columnwidth]{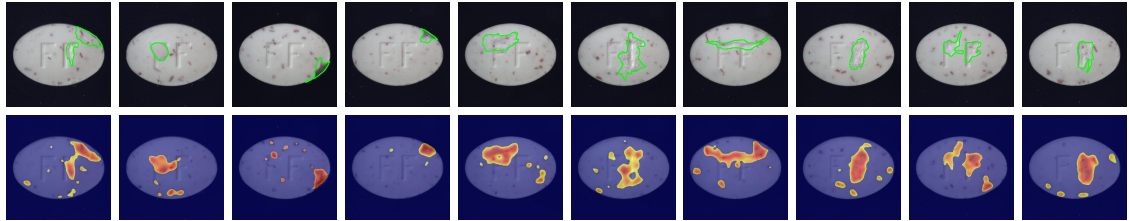}
	\caption{Visualization of segmentation results for the pill category on MVTec AD. The first row represents the input images, with green outlines indicating the ground truth regions. The second row shows the results of anomaly segmentation.}
	\label{fig_r11}
\end{figure*}

\begin{figure*}[tb]
	\centering
	\includegraphics[width=1.8\columnwidth]{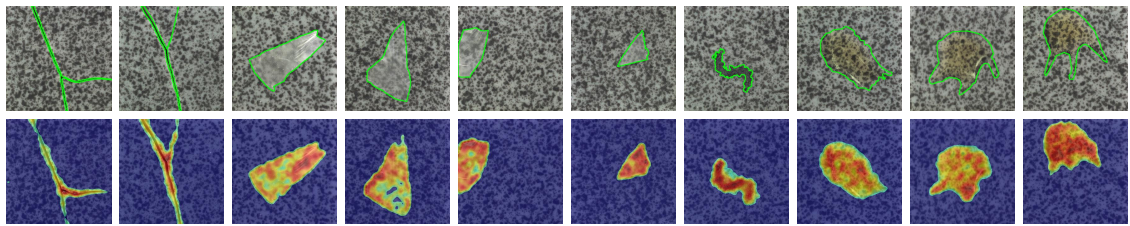}
	\caption{Visualization of segmentation results for the tile category on MVTec-AD. The first row represents the input images, with green outlines indicating the ground truth regions. The second row shows the results of anomaly segmentation.}
	\label{fig_r12}
\end{figure*}

\begin{figure*}[tb]
	\centering
	\includegraphics[width=1.8\columnwidth]{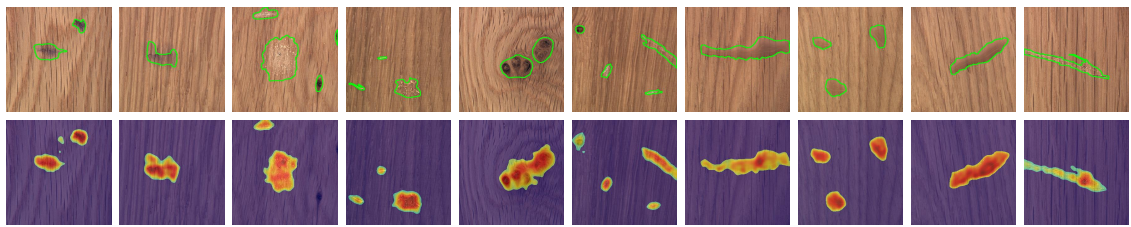}
	\caption{Visualization of segmentation results for the wood category on MVTec-AD. The first row represents the input images, with green outlines indicating the ground truth regions. The second row shows the results of anomaly segmentation.}
	\label{fig_r13}
\end{figure*}

\begin{figure*}[tb]
	\centering
	\includegraphics[width=1.8\columnwidth]{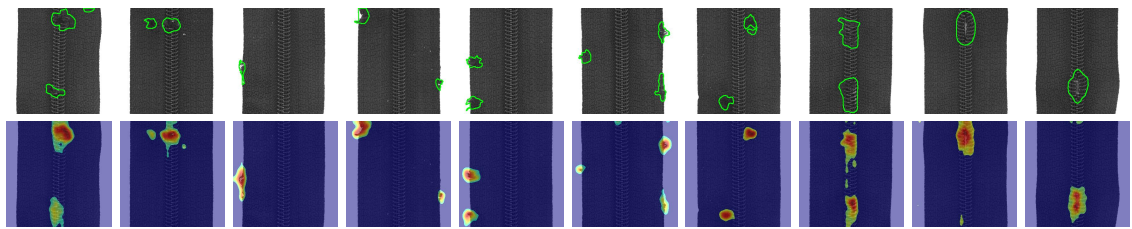}
	\caption{Visualization of segmentation results for the zipper category on MVTec-AD. The first row represents the input images, with green outlines indicating the ground truth regions. The second row shows the results of anomaly segmentation.}
	\label{fig_r14}
\end{figure*}

\begin{figure*}[tb]
	\centering
	\includegraphics[width=1.8\columnwidth]{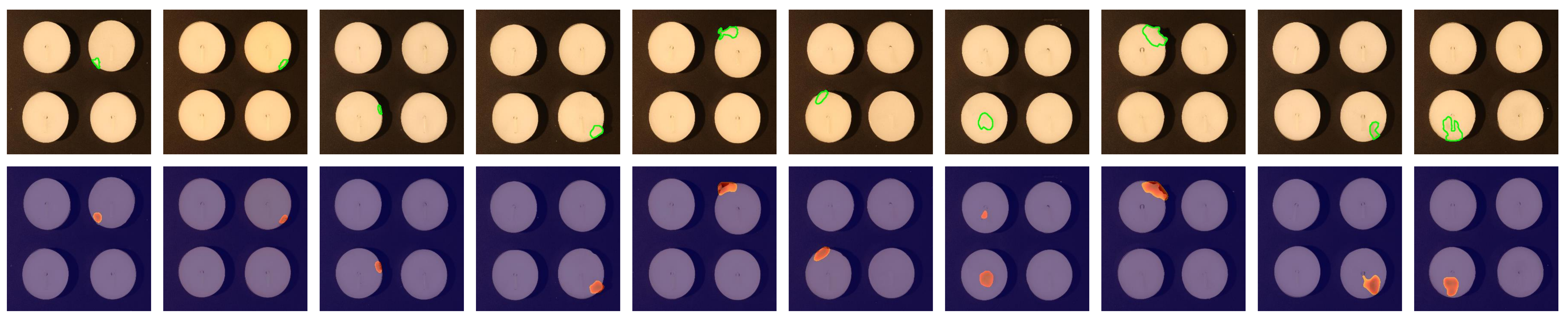}
	\caption{Visualization of segmentation results for the candle category on VisA. The first row represents the input images, with green outlines indicating the ground truth regions. The second row shows the results of anomaly segmentation.}
	\label{fig_r15}
\end{figure*}

\begin{figure*}[tb]
	\centering
	\includegraphics[width=1.8\columnwidth]{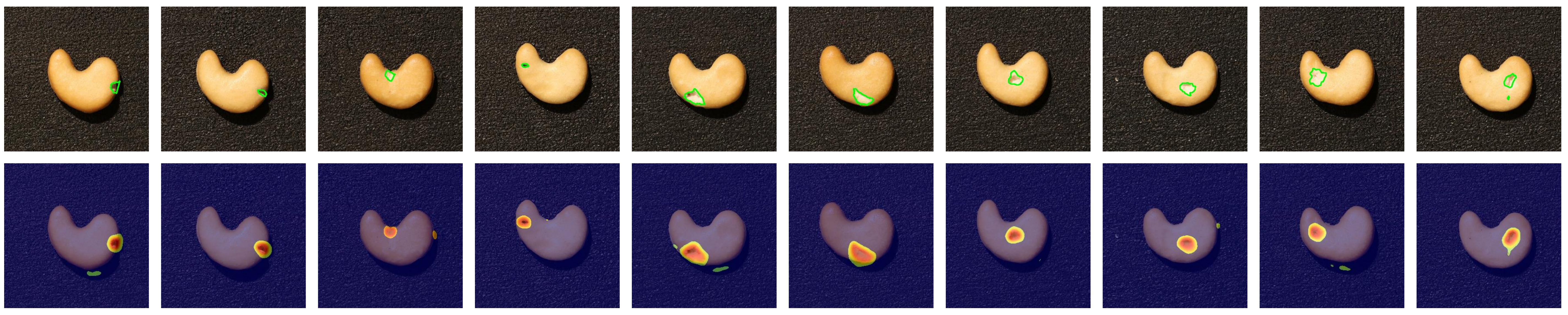}
	\caption{Visualization of segmentation results for the cashew category on VisA. The first row represents the input images, with green outlines indicating the ground truth regions. The second row shows the results of anomaly segmentation.}
	\label{fig_r16}
\end{figure*}

\begin{figure*}[tb]
	\centering
	\includegraphics[width=1.8\columnwidth]{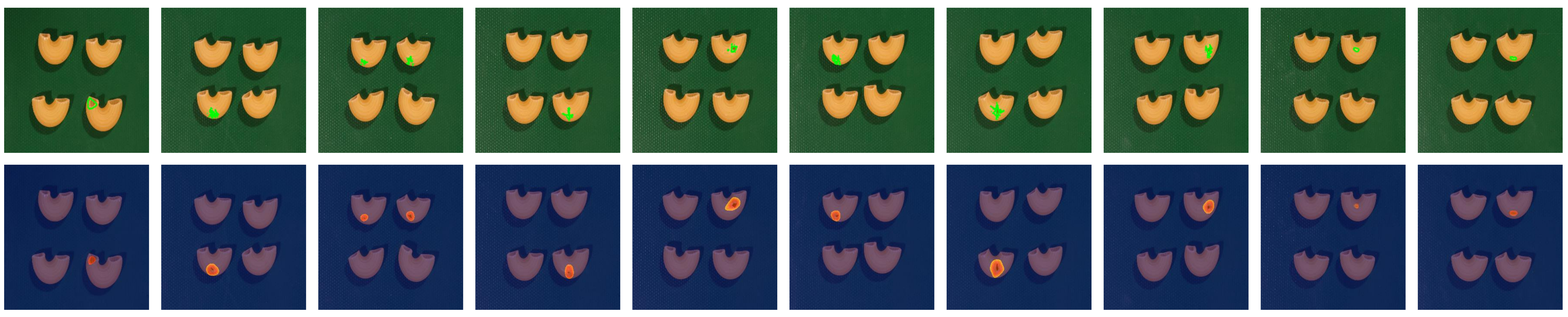}
	\caption{Visualization of segmentation results for the macaroni1 category on VisA. The first row represents the input images, with green outlines indicating the ground truth regions. The second row shows the results of anomaly segmentation.}
	\label{fig_r17}
\end{figure*}

\begin{figure*}[tb]
	\centering
	\includegraphics[width=1.8\columnwidth]{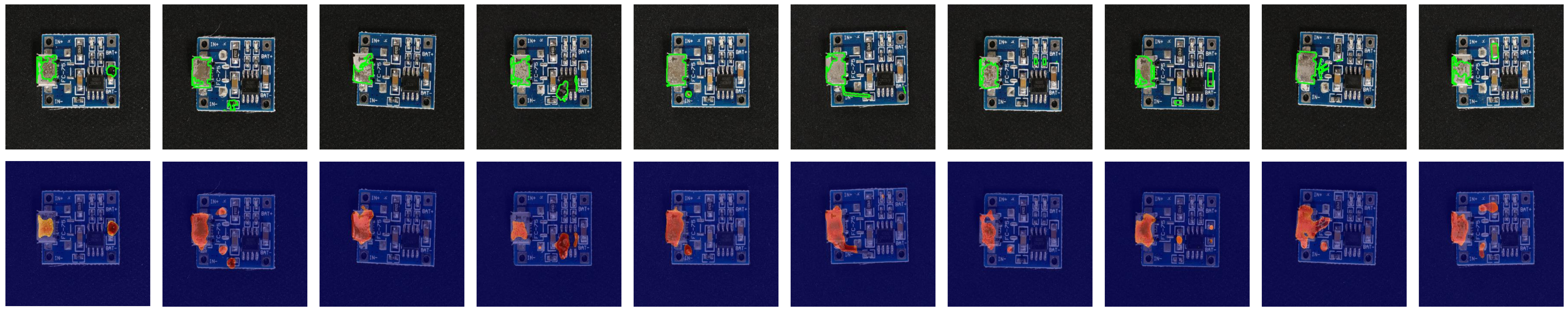}
	\caption{Visualization of segmentation results for the pcb4 category on VisA. The first row represents the input images, with green outlines indicating the ground truth regions. The second row shows the results of anomaly segmentation.}
	\label{fig_r18}
\end{figure*}

\begin{figure*}[tb]
	\centering
	\includegraphics[width=1.8\columnwidth]{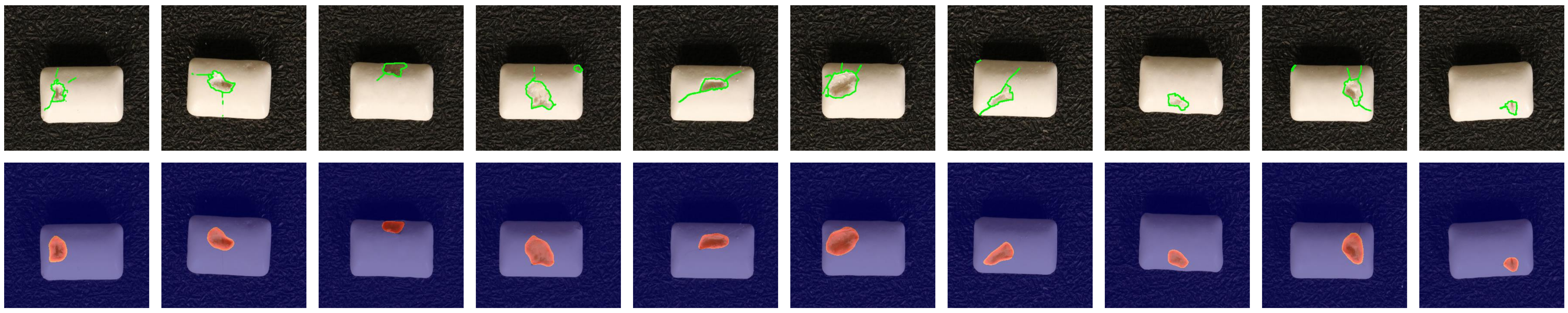}
	\caption{Visualization of segmentation results for the chewinggum category on VisA. The first row represents the input images, with green outlines indicating the ground truth regions. The second row shows the results of anomaly segmentation.}
	\label{fig_r19}
\end{figure*}

\begin{figure*}[tb]
	\centering
	\includegraphics[width=1.8\columnwidth]{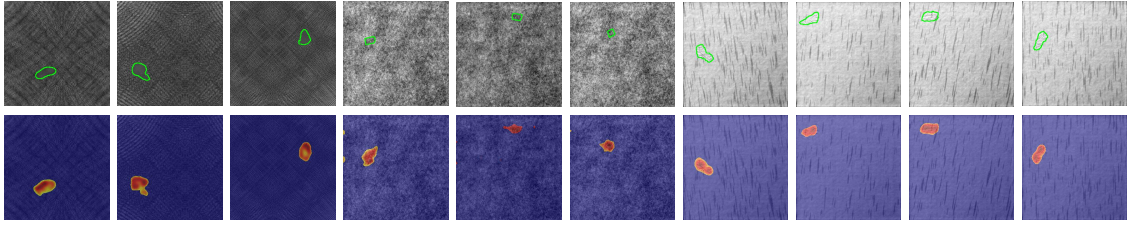}
	\caption{Visualization of segmentation results on DAGM. The first row represents the input images, with green outlines indicating the ground truth regions. The second row shows the results of anomaly segmentation.}
	\label{fig_r20}
\end{figure*}

\begin{figure*}[tb]
	\centering
	\includegraphics[width=1.8\columnwidth]{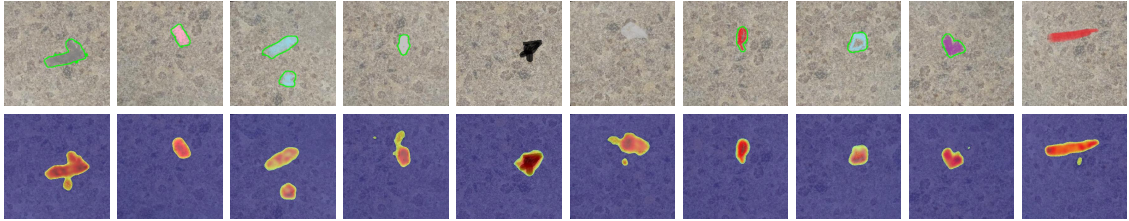}
	\caption{Visualization of segmentation results for the blotchy category on DTD-Synthetic. The first row represents the input images, with green outlines indicating the ground truth regions. The second row shows the results of anomaly segmentation.}
	\label{fig_r21}
\end{figure*}

\begin{figure*}[tb]
	\centering
	\includegraphics[width=1.8\columnwidth]{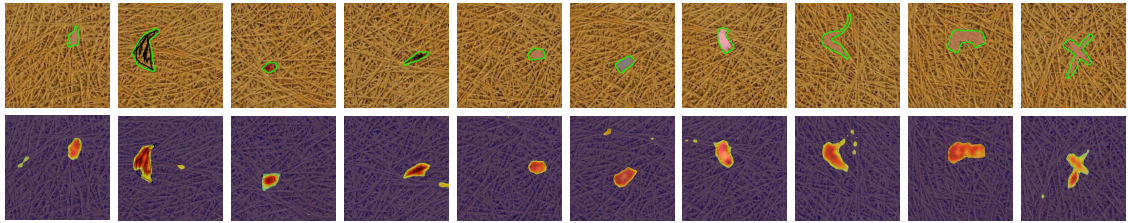}
	\caption{Visualization of segmentation results for the fibrous category on DTD-Synthetic. The first row represents the input images, with green outlines indicating the ground truth regions. The second row shows the results of anomaly segmentation.}
	\label{fig_r22}
\end{figure*}

\begin{figure*}[tb]
	\centering
	\includegraphics[width=1.8\columnwidth]{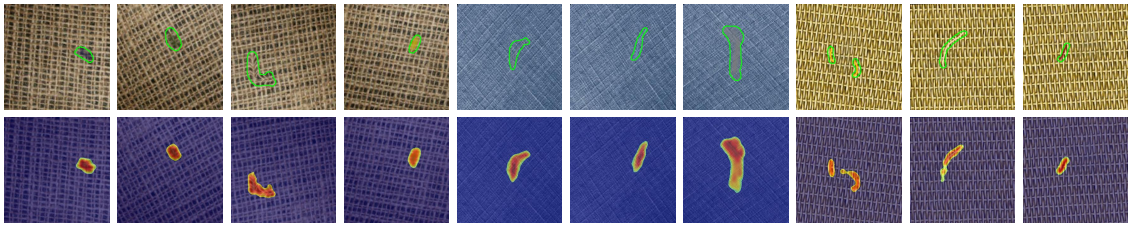}
	\caption{Visualization of segmentation results for the woven category on DTD-Synthetic. The first row represents the input images, with green outlines indicating the ground truth regions. The second row shows the results of anomaly segmentation.}
	\label{fig_r23}
\end{figure*}

\begin{figure*}[tb]
	\centering
	\includegraphics[width=1.8\columnwidth]{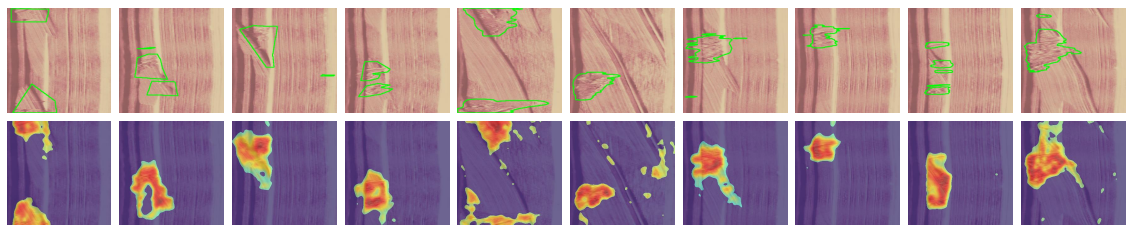}
	\caption{Visualization of segmentation results for the wood category on BTAD. The first row represents the input images, with green outlines indicating the ground truth regions. The second row shows the results of anomaly segmentation.}
	\label{fig_r24}
\end{figure*}

\begin{figure*}[tb]
	\centering
	\includegraphics[width=1.8\columnwidth]{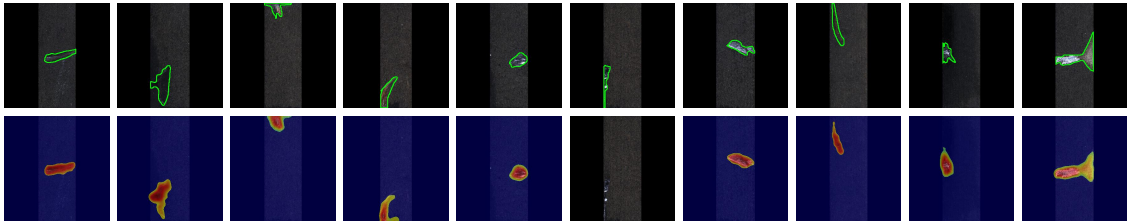}
	\caption{Visualization of segmentation results on KSDD2. The first row represents the input images, with green outlines indicating the ground truth regions. The second row shows the results of anomaly segmentation.}
	\label{fig_r25}
\end{figure*}

\begin{figure*}[tb]
	\centering
	\includegraphics[width=1.8\columnwidth]{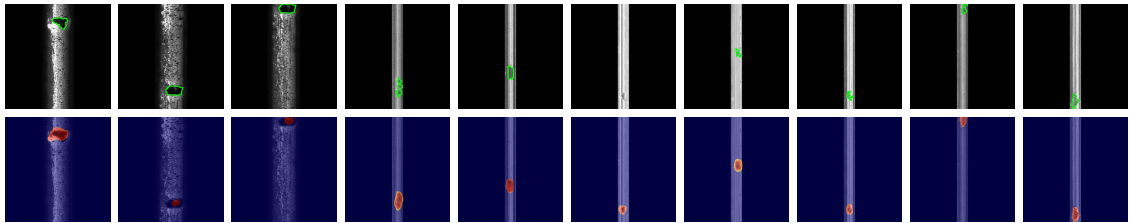}
	\caption{Visualization of segmentation results on RSDD. The first row represents the input images, with green outlines indicating the ground truth regions. The second row shows the results of anomaly segmentation.}
	\label{fig_r26}
\end{figure*}

\begin{figure*}[tb]
	\centering
	\includegraphics[width=1.8\columnwidth]{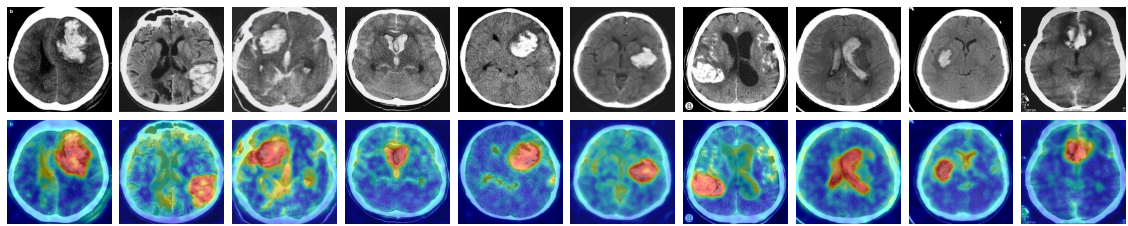}
	\caption{Visualization of segmentation results on HeadCT. Since there are no pixel-level annotations, this dataset is only used for anomaly classification. The first row represents the input image, while the second row shows the anomaly maps.}
	\label{fig_r27}
\end{figure*}

\begin{figure*}[tb]
	\centering
	\includegraphics[width=1.8\columnwidth]{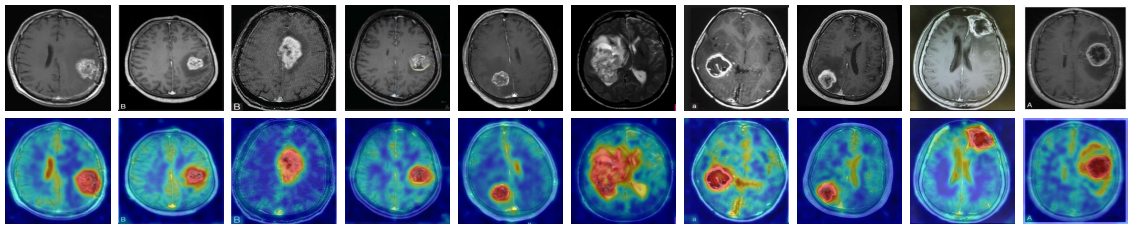}
	\caption{Visualization of segmentation results on BrainMRI. Since there are no pixel-level annotations, this dataset is only used for anomaly classification. The first row represents the input image, while the second row shows the anomaly maps.}
	\label{fig_r28}
\end{figure*}

\begin{figure*}[tb]
	\centering
	\includegraphics[width=1.8\columnwidth]{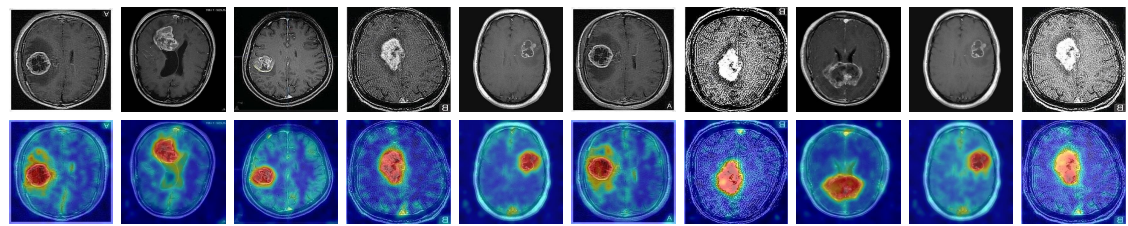}
	\caption{Visualization of segmentation results on Br35H. Since there are no pixel-level annotations, this dataset is only used for anomaly classification. The first row represents the input image, while the second row shows the anomaly maps.}
	\label{fig_r29}
\end{figure*}

\begin{figure*}[tb]
	\centering
	\includegraphics[width=1.8\columnwidth]{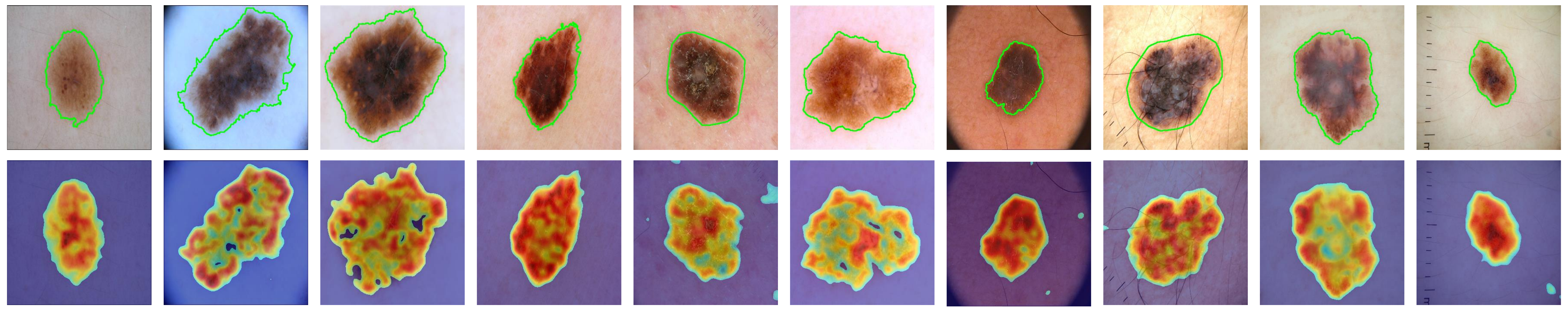}
	\caption{Visualization of segmentation results on ISIC. The first row represents the input images, with green outlines indicating the ground truth regions. The second row shows the results of anomaly segmentation.}
	\label{fig_r30}
\end{figure*}

\begin{figure*}[tb]
	\centering
	\includegraphics[width=1.8\columnwidth]{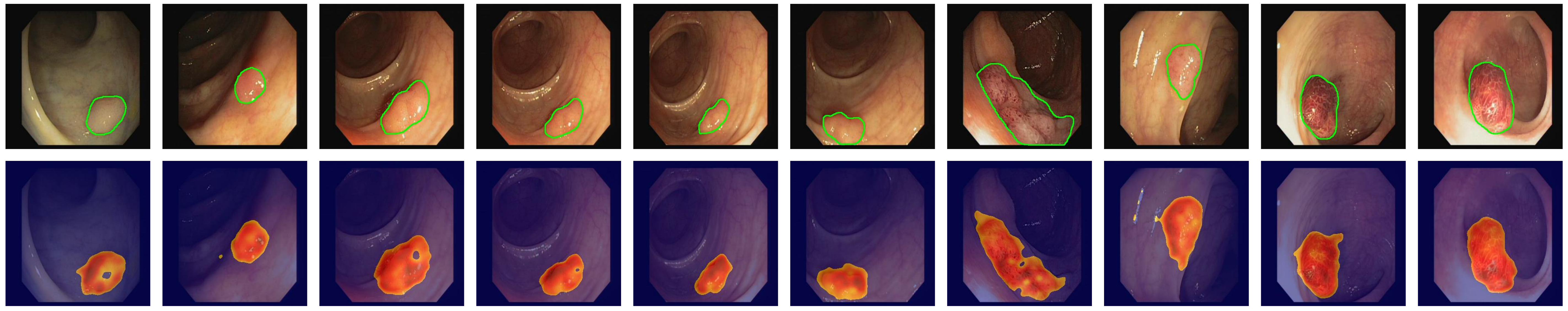}
	\caption{Visualization of segmentation results on CVC-ClinicDB. The first row represents the input images, with green outlines indicating the ground truth regions. The second row shows the results of anomaly segmentation.}
	\label{fig_r31}
\end{figure*}

\begin{figure*}[tb]
	\centering
	\includegraphics[width=1.8\columnwidth]{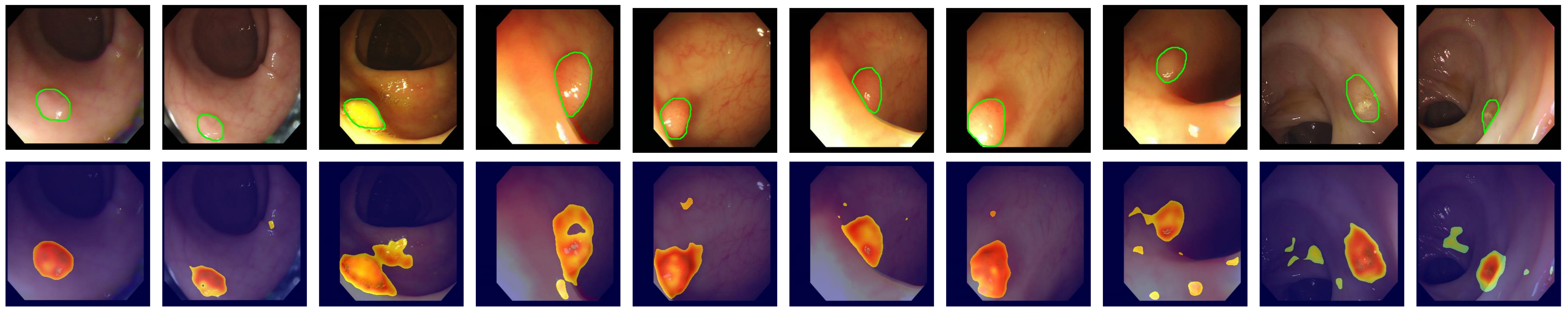}
	\caption{Visualization of segmentation results on CVC-ColonDB. The first row represents the input images, with green outlines indicating the ground truth regions. The second row shows the results of anomaly segmentation.}
	\label{fig_r32}
\end{figure*}

\begin{figure*}[tb]
	\centering
	\includegraphics[width=1.8\columnwidth]{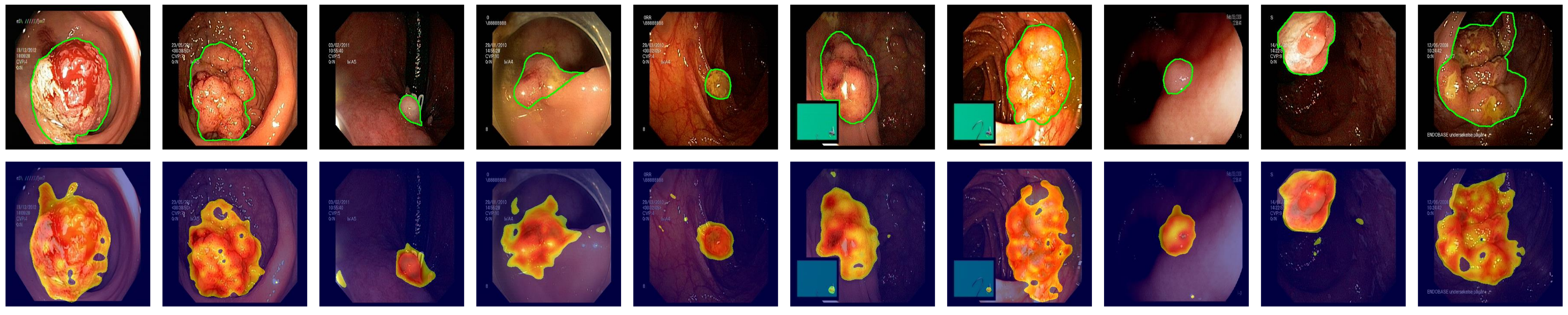}
	\caption{Visualization of segmentation results on Endo. The first row represents the input images, with green outlines indicating the ground truth regions. The second row shows the results of anomaly segmentation.}
	\label{fig_r33}
\end{figure*}

\begin{figure*}[tb]
	\centering
	\includegraphics[width=1.8\columnwidth]{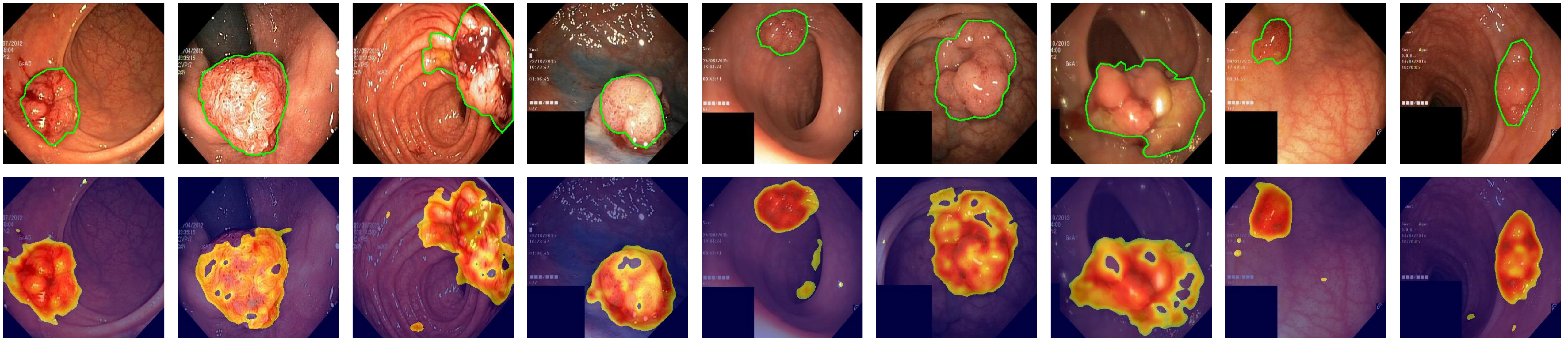}
	\caption{Visualization of segmentation results on kvasir. The first row represents the input images, with green outlines indicating the ground truth regions. The second row shows the results of anomaly segmentation.}
	\label{fig_r34}
\end{figure*}

In Tables \ref{tab_r4} to \ref{tab_r9}, we present the detailed quantitative results for each specific category of the MVTec-AD and VisA datasets. In Figures \ref{fig_r5} to \ref{fig_r34}, we provide more extensive qualitative results for various categories across different industrial and medical datasets. 
\section{Limitations}   \label{secE}
Our Bayes-PFL has already demonstrated the state-of-the-art ZSAD performance on 15 industrial and medical datasets. However, it still faces several limitations in practical applications: 1) During the inference stage, sampling from the learned prompt distribution incurs additional time overhead, hindering the model's ability to achieve optimal inference efficiency; 2) The vision-language model CLIP used in this study has limited capacity to understand textual information and extract fine-grained image features. In future work, we plan to investigate alternative vision-language models that may be better suited for fine-grained anomaly detection task.

\begin{table*}[t]
	\centering
	\caption{Comparison of different categories in terms of pixel-level AUROC. The best results are shown in \textbf{bold}.}
	\label{tab_r4}
	\renewcommand{\arraystretch}{1}
	\resizebox{1.4\columnwidth}{!}
	{
		\begin{tabular}{>{\centering\arraybackslash}p{2cm}>{\centering\arraybackslash}p{1.4cm}*{6}{>{\centering\arraybackslash}p{1.9cm}}}
			\toprule
			Dataset                    & Category    & WinCLIP & APRIL-GAN & CLIP-AD & AnomalyCLIP & AdaCLIP & Bayes-PFL \\  \midrule
			\multirow{16}{*}{MVTec-AD} & bottle      & 89.5    & 83.5      & 91.2    & 90.4        & 83.8    & \textbf{94.0}      \\
			& cable       & 77.0    & 72.2      & 76.2    & \textbf{78.9}        & 85.6    & 78.4      \\
			& capsule     & 86.9    & 92.0      & 95.1    & 95.8        & 86.2    & \textbf{96.4}      \\
			& carpet      & 95.4    & 98.4      & 99.1    & 98.8        & 94.8    & \textbf{99.6}      \\
			& grid        & 82.2    & 95.8      & 96.3    & 97.3        & 90.6    & \textbf{98.2}      \\
			& hazelnut    & 94.3    & 96.1      & 97.2    & 97.2        & 98.7    & \textbf{97.3}      \\
			& leather     & 96.7    & 99.1      & 99.3    & 98.6        & 97.8    & \textbf{99.6}      \\
			& metal\_nut  & 61.0    & 65.5      & 58.9    & 74.6        & 55.4    & \textbf{77.5}      \\
			& pill        & 80.0    & 76.2      & 83.7    & \textbf{91.8}        & 77.5    & 89.1      \\
			& screw       & 89.6    & 97.8      & 98.7    & 97.5        & \textbf{99.2}    & 98.4      \\
			& tile        & 77.6    & 92.7      & 94.5    & \textbf{94.7}        & 83.9    & 92.8      \\
			& toothbrush  & 86.9    & 95.8      & 92.7    & 91.9        & 93.4    & \textbf{93.7}      \\
			& transistor  & 74.7    & 62.4      & \textbf{75.5}    & 70.8        & 71.4    & 67.5      \\
			& wood        & 93.4    & 95.8      & 96.9    & 96.4        & 91.2    & \textbf{97.1}      \\
			& zipper      & 91.6    & 91.1      & 92.8    & 91.2        & 91.8    & \textbf{97.4}      \\   \cmidrule(lr){2-8}
			& mean        & 85.1    & 87.6      & 89.8    & 91.1        & 86.8    & \textbf{91.8}      \\   \midrule\midrule
			\multirow{13}{*}{VisA}     & candle      & 88.9    & 97.8      & 98.7    & 98.8        & 98.6    & \textbf{99.2}      \\
			& capsules    & 81.6    & 97.5      & 97.4    & 94.9        & 96.1    & \textbf{98.7}      \\
			& cashew      & 84.7    & 86.0      & 91.4    & 93.7        & 97.2    & \textbf{91.6}      \\
			& chewinggum  & 93.3    & 99.5      & 99.2    & 99.2        & 99.2    & \textbf{99.6}      \\
			& fryum       & 88.5    & 92.0      & 93.0    & 94.6        & 93.6    & \textbf{94.7}      \\
			& macaroni1   & 70.9    & 98.8      & 98.7    & 98.3        & 98.8    & \textbf{99.3}      \\
			& macaroni2   & 59.3    & 97.8      & 97.6    & 97.6        & 98.2    & \textbf{98.3}      \\
			& pcb1        & 61.2    & 92.7      & 92.6    & 94.0        & 90.7    & \textbf{93.8}      \\
			& pcb2        & 71.6    & 89.8      & 91.0    & 92.4        & 91.3    & \textbf{92.4}      \\
			& pcb3        & \textbf{85.3}    & 88.4      & 87.5    & 88.3        & 87.7    & 88.4      \\
			& pcb4        & 94.4    & 94.6      & 95.9    & \textbf{95.7}        & 94.6    & 95.3      \\
			& pipe\_fryum & 75.4    & 96.0      & 96.9    & \textbf{98.2}        & 95.7    & 95.4      \\  \cmidrule(lr){2-8}
			& mean        & 79.6    & 94.2      & 95.0    & 95.5        & 95.1    & \textbf{95.6}       \\ \bottomrule
		\end{tabular}
	}
\end{table*}

% \usepackage{multirow}
\begin{table*}[t]
	\centering
	\caption{Comparison of different categories in terms of image-level AUROC. The best results are shown in \textbf{bold}.}
	\label{tab_r5}
	\renewcommand{\arraystretch}{1}
	\resizebox{1.4\columnwidth}{!}
	{
		\begin{tabular}{>{\centering\arraybackslash}p{2cm}>{\centering\arraybackslash}p{1.4cm}*{6}{>{\centering\arraybackslash}p{1.9cm}}}
			\toprule
			Dataset                    & Category    & WinCLIP & APRIL-GAN & CLIP-AD & AnomalyCLIP & AdaCLIP & Bayes-PFL \\  \midrule
			\multirow{16}{*}{MVTec-AD} & bottle      & \textbf{99.2}    & 92.0      & 96.4    & 88.7        & 95.6    & 95.6      \\
			& cable       & \textbf{86.5}    & 88.2      & 80.4    & 70.3        & 79.0    & 81.5      \\
			& capsule     & 72.9    & 79.8      & 82.8    & 89.5        & 89.3    & \textbf{92.4}      \\
			& carpet      & 100.0   & 99.4      & 99.5    & 99.9        & 100.0   & \textbf{100}       \\
			& grid        & 98.8    & 86.2      & 94.1    & 97.8        & 99.2    & \textbf{99.7}      \\
			& hazelnut    & 93.9    & 89.4      & 98.0    & 97.2        & 95.5    & \textbf{95.9}      \\
			& leather     & 100.0   & 99.7      & 100.0   & 99.8        & 100.0   & \textbf{100}       \\
			& metal\_nut  & \textbf{97.1}    & 68.2      & 75.1    & 92.4        & 79.9    & 75.9      \\
			& pill        & 79.1    & 80.8      & 87.7    & 81.1        & \textbf{92.6}    & 82.2      \\
			& screw       & 83.3    & 85.1      & 89.1    & 82.1        & 83.9    & \textbf{89.4}      \\
			& tile        & \textbf{100.0}   & 99.8      & 99.6    & 100         & 99.7    & 99.3      \\
			& toothbrush  & 87.5    & 53.2      & 76.1    & 85.3        & \textbf{95.2}    & 90.5      \\
			& transistor  & 88.0    & 80.9      & 79.3    & \textbf{93.9}        & 82      & 84.5      \\
			& wood        & \textbf{99.4}    & 98.9      & 98.9    & 96.9        & 98.5    & 98.1      \\
			& zipper      & 91.5    & 89.4      & 88.6    & 98.4        & 89.4    & \textbf{99.7}      \\  \cmidrule(lr){2-8}
			& mean        & 91.8    & 86.1      & 89.8    & 91.5        & 92.0      & \textbf{92.3}      \\  \midrule\midrule
			\multirow{13}{*}{VisA}     & candle      & 95.4    & 82.5      & 89.4    & 80.9        & \textbf{95.9}    & 92.3      \\
			& capsules    & 85.0    & 62.3      & 75.2    & 82.7        & 81.1    & \textbf{92.1}      \\
			& cashew      & \textbf{92.1}    & 86.7      & 83.7    & 76.0        & 89.6    & 91.3      \\
			& chewinggum  & 96.5    & 96.5      & 95.6    & 97.2        & \textbf{98.5}    & 97.5      \\
			& fryum       & 80.3    & 93.8      & 78.7    & 92.7        & 89.5    & \textbf{95.8}      \\
			& macaroni1   & 76.2    & 69.5      & 80.0    & 86.7        & 86.3    & \textbf{90.8}      \\
			& macaroni2   & 63.7    & 65.7      & 67.0    & 72.2        & 56.7    & \textbf{67.6}      \\
			& pcb1        & 73.6    & 50.6      & 68.6    & 85.2        & \textbf{74.0}    & 66.1      \\
			& pcb2        & 51.2    & 71.6      & 69.7    & 62.0        & 71.1    & \textbf{76.5}      \\
			& pcb3        & 73.4    & 66.9      & 67.3    & 61.7        & 75.2    & \textbf{79.0}      \\
			& pcb4        & 79.6    & 94.6      & 96.2    & 93.9        & 89.6    & \textbf{96.3}      \\
			& pipe\_fryum & 69.7    & 89.4      & 86.5    & 92.3        & 88.8    & \textbf{98.7}      \\ \cmidrule(lr){2-8}
			& mean        & 78.1    & 78.0      & 79.8    & 82.1        & 83.0    & \textbf{87.0}     \\ \bottomrule
		\end{tabular}
	}
\end{table*}

% \usepackage{multirow}
\begin{table*}[t]
	\centering
	\caption{Comparison of different categories in terms of pixel-level AP. The best results are shown in \textbf{bold}.}
	\label{tab_r6}
	\renewcommand{\arraystretch}{1}
	\resizebox{1.4\columnwidth}{!}
	{
		\begin{tabular}{>{\centering\arraybackslash}p{2cm}>{\centering\arraybackslash}p{1.4cm}*{6}{>{\centering\arraybackslash}p{1.9cm}}}
			\toprule
			Dataset                    & Category    & WinCLIP & APRIL-GAN & CLIP-AD & AnomalyCLIP & AdaCLIP & Bayes-PFL \\  \midrule
			\multirow{16}{*}{MVTec-AD} & bottle      & 49.8    & 53.0      & 56.8    & 55.3        & 49.8    & \textbf{67.4}      \\
			& cable       & 6.2     & 18.2      & 17.3    & 12.3        & 16.5    & \textbf{20.7}      \\
			& capsule     & 8.6     & 29.6      & 27.2    & 27.7        & 24.8    & \textbf{32.6}      \\
			& carpet      & 25.9    & 67.5      & 65.4    & 56.6        & 63.5    & \textbf{84.5}      \\
			& grid        & 5.7     & 36.5      & 30.7    & 24.1        & 27.8    & \textbf{41.3}      \\
			& hazelnut    & 33.3    & 49.7      & 59.2    & 43.4        & \textbf{69.5}    & 56.9      \\
			& leather     & 20.4    & 52.3      & 50.5    & 22.7        & 53.6    & \textbf{63.7}      \\
			& metal\_nut  & 10.8    & 25.9      & 21.2    & 26.4        & 19.9    & \textbf{27.7}      \\
			& pill        & 7.0     & 23.6      & 26.1    & \textbf{34.1}        & 25.8    & 30.5      \\
			& screw       & 5.4     & 33.7      & 39.1    & 27.5        & \textbf{41.6}    & 40.0      \\
			& tile        & 21.2    & 66.3      & 65.2    & 61.7        & 48.8    & \textbf{76.6}      \\
			& toothbrush  & 5.5     & \textbf{43.2}      & 29.9    & 19.3        & 24.7    & 29.4      \\
			& transistor  & \textbf{20.2}    & 11.7      & 14.2    & 15.6        & 11.9    & 13.3      \\
			& wood        & 32.9    & 61.8      & 59.4    & 52.6        & 56.6    & \textbf{73.2}      \\
			& zipper      & 19.4    & 38.7      & 38.5    & 38.7        & 36.0    & \textbf{66.5}      \\   \cmidrule(lr){2-8}
			& mean        & 18.0    & 40.8      & 40.0    & 34.5        & 38.1    & \textbf{48.3}      \\  \midrule\midrule
			\multirow{13}{*}{VisA}     & candle      & 2.4     & 29.9      & 36.6    & 25.6        & 45.3    & \textbf{38.5}      \\
			& capsules    & 1.4     & 40.0      & 38.5    & 29.3        & 18.2    & \textbf{47.2}      \\
			& cashew      & 4.8     & 15.1      & 24.1    & 19.6        & \textbf{44.8}    & 25        \\
			& chewinggum  & 24.0    & 83.6      & 83.4    & 56.3        & \textbf{87.6}    & 83.8      \\
			& fryum       & 11.1    & 22.1      & 22.4    & 22.6        & 24.0    & \textbf{28.5}      \\
			& macaroni1   & 0.03    & 24.8      & 23.2    & 14.9        & \textbf{27.1}    & 23.9      \\
			& macaroni2   & 0.02    & \textbf{6.8}       & 2.3     & 1.5         & 3.0     & 4.0         \\
			& pcb1        & 0.4     & 8.4       & 7.2     & \textbf{8.6}         & 7.8     & 7.6       \\
			& pcb2        & 0.4     & 15.4      & 8.2     & 9.1         & 17.5    & \textbf{17.9}      \\
			& pcb3        & 0.7     & 14.1      & 11.7    & 4.3         & 16.1    & \textbf{19}        \\
			& pcb4        & 15.5    & 24.9      & 31.2    & 30.6        & 34.2    & \textbf{31.9}      \\
			& pipe\_fryum & 4.4     & 23.6      & 27.2    & \textbf{33.2}        & 24.4    & 31.2      \\  \cmidrule(lr){2-8}
			& mean        & 5.0     & 25.7      & 26.3    & 21.3        & 29.2    & \textbf{29.8}      \\ \bottomrule
		\end{tabular}
	}
\end{table*}

\begin{table*}[t]
	\centering
	\caption{Comparison of different categories in terms of image-level AP. The best results are shown in \textbf{bold}.}
	\label{tab_r7}
	\renewcommand{\arraystretch}{1}
	\resizebox{1.4\columnwidth}{!}
	{
		\begin{tabular}{>{\centering\arraybackslash}p{2cm}>{\centering\arraybackslash}p{1.4cm}*{6}{>{\centering\arraybackslash}p{1.9cm}}}
			\toprule
			Dataset                    & Category    & WinCLIP & APRIL-GAN & CLIP-AD & AnomalyCLIP & AdaCLIP & Bayes-PFL \\  \midrule
			\multirow{16}{*}{MVTec-AD} & bottle      & 98.3    & 97.7      & 98.8    & 96.8        & 98.6    & \textbf{98.7}      \\
			& cable       & 86.2    & \textbf{92.9}      & 88.9    & 81.7        & 87.3    & 90.1      \\
			& capsule     & 93.4    & 95.4      & 96.4    & 97.8        & 97.8    & \textbf{98.4}      \\
			& carpet      & 99.9    & 99.8      & 99.8    & 99.9        & 100.0   & \textbf{100}       \\
			& grid        & 99.8    & 94.9      & 97.9    & 99.3        & 99.7    & \textbf{99.9}      \\
			& hazelnut    & 96.3    & 94.6      & \textbf{99.0}    & 98.5        & 97.5    & 98.0      \\
			& leather     & 100.0   & 99.9      & 100.0   & 99.9        & 100.0   & \textbf{100}       \\
			& metal\_nut  & 97.9    & 91.8      & 94.4    & \textbf{98.1}        & 95.6    & 94.6      \\
			& pill        & \textbf{96.5}    & 96.1      & 97.6    & 95.3        & 98.6    & 95.9      \\
			& screw       & 88.4    & 93.6      & 96.2    & 92.9        & 93.0    & \textbf{96.0}      \\
			& tile        & 99.9    & 99.9      & 99.8    & \textbf{100}         & 99.9    & 99.7      \\
			& toothbrush  & 96.7    & 71.9      & 90.2    & 93.9        & 97.9    & \textbf{96.8}      \\
			& transistor  & 74.9    & 77.6      & 73.7    & \textbf{92.1}        & 83.8    & 82.3      \\
			& wood        & 98.8    & 99.6      & 99.6    & 99.2        & 99.5    &\textbf{ 99.7}      \\
			& zipper      & 98.9    & 97.1      & 96.9    & 99.5        & 97.1    & \textbf{99.9}      \\ \cmidrule(lr){2-8}
			& mean        & 95.1    & 93.5      & 95.3    & 96.2        & 96.4    & \textbf{96.7}      \\  \midrule\midrule
			\multirow{13}{*}{VisA}     & candle      & 95.6    & 85.9      & 91.6    & 82.6        & \textbf{96.4}    & 93.8      \\
			& capsules    & 80.9    & 74.6      & 86.6    & 89.4        & 86.7    & \textbf{95.9}      \\
			& cashew      & 95.2    & 93.9      & 92.4    & 89.3        & 95.4    & \textbf{96.3}      \\
			& chewinggum  & 98.8    & 98.4      & 98.1    & 98.8        & 99.4    & \textbf{99.5}      \\
			& fryum       & 92.5    & 97.0      & 90.4    & 96.6        & 95.1    & \textbf{98.4}      \\
			& macaroni1   & 64.5    & 67.5      & 81.1    & 85.5        & 85.0    & \textbf{92.2}      \\
			& macaroni2   & 65.2    & 64.9      & 65.3    & \textbf{70.8}        & 54.3    & 70.4      \\
			& pcb1        & 74.6    & 54.6      & 72.5    & \textbf{86.7}        & 73.5    & 70.0      \\
			& pcb2        & 44.2    & 73.8      & 71.4    & 64.4        & 71.6    & \textbf{77.3}      \\
			& pcb3        & 66.2    & 70.5      & 71.9    & 69.4        & 77.9    & \textbf{81.1}      \\
			& pcb4        & 70.1    & 94.8      & 96.0    & 94.3        & 89.8    & \textbf{96.1}      \\
			& pipe\_fryum & 82.1    & 94.6      & 93.7    & 96.3        & 93.9    & \textbf{99.4}      \\   \cmidrule(lr){2-8}
			& mean        & 77.5    & 81.4      & 84.3    & 85.4        & 84.9    & \textbf{89.2}     \\ \bottomrule
		\end{tabular}
	}
\end{table*}

% \usepackage{multirow}
\begin{table*}[t]
	\centering
	\caption{Comparison of different categories in terms of pixel-level PRO. The best results are shown in \textbf{bold}.}
	\label{tab_r8}
	\renewcommand{\arraystretch}{1}
	\resizebox{1.4\columnwidth}{!}
	{
		\begin{tabular}{>{\centering\arraybackslash}p{2cm}>{\centering\arraybackslash}p{1.4cm}*{6}{>{\centering\arraybackslash}p{1.9cm}}}
			\toprule
			Dataset                    & Category    & WinCLIP & APRIL-GAN & CLIP-AD & AnomalyCLIP & AdaCLIP & Bayes-PFL \\  \midrule
			\multirow{16}{*}{MVTec-AD} & bottle      & 76.4    & 45.6      & 71.7    & 80.8        & 26.9    & \textbf{87.6}      \\
			& cable       & 42.9    & 25.7      & 51.4    & 64.0        & 15.2    & \textbf{69.1}      \\
			& capsule     & 62.1    & 51.3      & 62.6    & 87.6        & 65.7    & \textbf{93.4}      \\
			& carpet      & 84.1    & 48.5      & 83.0    & 90.0        & 19.6    & \textbf{98.6}      \\
			& grid        & 57.0    & 31.6      & 67.8    & 75.4        & 46.2    & \textbf{93.5}      \\
			& hazelnut    & 81.6    & 70.3      & 83.5    & \textbf{92.5}        & 42.3    & 87.1      \\
			& leather     & 91.1    & 72.4      & 95.5    & 92.2        & 55.9    & \textbf{99.1}      \\
			& metal\_nut  & 31.8    & 38.4      & 72.7    & 71.1        & 20.7    & \textbf{74.9}      \\
			& pill        & 65.0    & 65.4      & 87.5    & 88.1        & 37.0    & \textbf{93.8}      \\
			& screw       & 68.5    & 67.1      & 88.5    & 88.0        & 75.3    & \textbf{93.0}      \\
			& tile        & 51.2    & 26.7      & 61.1    & 87.4        & 7.7     & \textbf{90.3}      \\
			& toothbrush  & 67.7    & 54.5      & 83.5    & \textbf{88.5}        & 25.6    & 87.6      \\
			& transistor  & 43.4    & 21.3      & 34.9    & \textbf{58.2}        & 6.7     & 55.5      \\
			& wood        & 74.1    & 31.1      & 85.6    & 91.5        & 58.3    & \textbf{95.1}      \\
			& zipper      & 71.7    & 10.7      & 30.2    & 65.4        & 3.4     & \textbf{90.1}      \\  \cmidrule(lr){2-8}
			& mean        & 64.6    & 44.0      & 70.6    & 81.4        & 33.8    & \textbf{87.4}      \\   \midrule\midrule
			\multirow{13}{*}{VisA}     & candle      & 83.5    & 92.5      & 94.4    & \textbf{96.5}        & 71.6    & 95.6      \\
			& capsules    & 35.3    & 86.7      & 87.2    & 78.9        & 80.3    & \textbf{87.4}      \\
			& cashew      & 76.4    & 91.7      & 90.6    & \textbf{91.9}        & 45.6    & 90.5      \\
			& chewinggum  & 70.4    & 87.3      & 82.7    & \textbf{90.9}        & 53.9    & 87.4      \\
			& fryum       & 77.4    & 89.7      & 87.5    & 86.8        & 55.6    & \textbf{90.4}      \\
			& macaroni1   & 34.3    & 93.2      & 93.8    & 89.7        & 86.6    & \textbf{96.5}      \\
			& macaroni2   & 21.4    & 82.3      & 83.9    & 83.9        & 84.8    & \textbf{89.3}      \\
			& pcb1        & 26.3    & 87.5      & 84.1    & 80.7        & 52.3    & \textbf{89.3}      \\
			& pcb2        & 37.2    & 75.6      & 77.7    & 78.2        & 77.5    & \textbf{78.3}      \\
			& pcb3        & 56.1    & 77.8      & 78.8    & 76.8        & 76.2    & \textbf{79.3}      \\
			& pcb4        & 80.4    & 86.8      & 88.9    & \textbf{89.4}        & 84.3    & 88.6      \\
			& pipe\_fryum & 82.3    & 90.9      & 93.2    & \textbf{96.1}        & 86.3    & 95.3      \\  \cmidrule(lr){2-8}
			& mean        & 56.8    & 86.8      & 86.9    & 87.0        & 71.3    & \textbf{88.9}      \\ \bottomrule
		\end{tabular}
	}
\end{table*}

% \usepackage{multirow}
\begin{table*}[t]
	\centering
	\caption{Comparison of different categories in terms of image-level F1-max. The best results are shown in \textbf{bold}.}
	\label{tab_r9}
	\renewcommand{\arraystretch}{1}
	\resizebox{1.4\columnwidth}{!}
	{
		\begin{tabular}{>{\centering\arraybackslash}p{2cm}>{\centering\arraybackslash}p{1.4cm}*{6}{>{\centering\arraybackslash}p{1.9cm}}}
			\toprule
			Dataset                    & Category    & WinCLIP & APRIL-GAN & CLIP-AD & AnomalyCLIP & AdaCLIP & Bayes-PFL \\   \midrule
			\multirow{16}{*}{MVTec-AD} & bottle      & \textbf{97.6}    & 92.8      & 94.6    & 90.9        & 93.7    & 94.6      \\
			& cable       & \textbf{84.5}    & 84.5      & 79.3    & 77.4        & 79.2    & 79.8      \\
			& capsule     & 91.4    & 92.0      & 91.1    & 91.7        & 91.8    & \textbf{95.2}      \\
			& carpet      & 99.4    & 98.3      & 99.4    & 99.4        & 100     & \textbf{100}       \\
			& grid        & 98.2    & 89.1      & 92.3    & 97.3        & 98.2    & \textbf{99.1}      \\
			& hazelnut    & 89.7    & 87.0      & \textbf{95.6}    & 92.7        & 93.7    & 92.3      \\
			& leather     & 100.0   & 98.9      & 100.0   & 99.5        & 100     & \textbf{100}       \\
			& metal\_nut  & \textbf{96.3}    & 89.4      & 89.4    & 93.6        & 89.4    & 89.4      \\
			& pill        & 91.6    & 91.6      & 92.1    & 92.1        & 93.7    & \textbf{93.9}      \\
			& screw       & 87.4    & 88.9      & 90.3    & 88.3        & 89.2    & \textbf{90.8 }     \\
			& tile        & \textbf{99.4}    & 98.8      & 98.8    & 100.0       & 98.8    & 98.3      \\
			& toothbrush  & 87.9    & 83.3      & 84.8    & \textbf{90.0}        & 96.7    & 88.9      \\
			& transistor  & 79.5    & 76.1      & 69.6    &\textbf{ 83.7}        & 77.1    & 78.4      \\
			& wood        & \textbf{98.3}    & 96.8      & 96.7    & 96.6        & 96.7    & 96.8      \\
			& zipper      & 92.9    & 90.8      & 90.4    & 97.9        & 91.1    & \textbf{99.2}      \\  \cmidrule(lr){2-8}
			& mean        & 92.9    & 90.4      & 91.1    & 92.8        & 92.7    & \textbf{93.1}      \\  \midrule\midrule
			\multirow{13}{*}{VisA}     & candle      & 89.4    & 76.9      & 82.2    & 75.6        & \textbf{90.2}    & 85.4      \\
			& capsules    & 83.9    & 78.1      & 77.9    & 82.2        & 82.8    & \textbf{89.1}      \\
			& cashew      & 88.4    & 85.7      & 83.7    & 80.3        & 87.5    & \textbf{89.4}      \\
			& chewinggum  & 94.8    & 93.2      & 92.8    & 94.8        & 96.0    & \textbf{96.4}      \\
			& fryum       & 82.7    & 91.8      & 81.4    & 90.1        & 87.0    & \textbf{93.9}      \\
			& macaroni1   & 74.2    & 70.8      & 77.4    & 80.4        & 80.0    & \textbf{82.1}      \\
			& macaroni2   & 69.8    & 69.3      & 69.7    & 71.2        & 67.8    & \textbf{69.9}      \\
			& pcb1        & 71.0    & 66.9      & 68.5    & \textbf{78.8}        & 72.1    & 66.9      \\
			& pcb2        & 67.1    & 69.7      & 70.9    & 67.8        & 72.5    & \textbf{73.3}      \\
			& pcb3        & 71.0    & 66.7      & 68.1    & 66.4        & 71.9    & \textbf{75.7}      \\
			& pcb4        & 74.9    & 87.3      & 91.3    & 87.8        & 82.1    & \textbf{91.4}      \\
			& pipe\_fryum & 80.7    & 88.1      & 86.6    & 89.8        & 88.5    & \textbf{95.4}      \\    \cmidrule(lr){2-8}
			& mean        & 79.0    & 78.7      & 79.2    & 80.4        & 81.6    & \textbf{84.1}       \\ \bottomrule
		\end{tabular}
	}
\end{table*}








\clearpage
\clearpage
%\section{Limitations}